\documentclass[times,twocolumn,final]{elsarticle} 
\usepackage[utf8]{inputenc} 
\usepackage[T1]{fontenc}
\usepackage{amsmath}
\usepackage{booktabs}
\usepackage{medima}
\usepackage{framed,multirow}

\usepackage{amssymb}
\usepackage{latexsym}
\usepackage{algorithm}
\usepackage{algpseudocode}
\usepackage{tabularx}
\usepackage{url}
\usepackage{makecell} 
\usepackage{hyperref}
\usepackage[table,xcdraw]{xcolor}
\usepackage{diagbox} 
\usepackage{arydshln}
\usepackage{threeparttable}
\usepackage{pifont}
\usepackage{array}
\usepackage{soul} 
\journal{Medical Image Analysis}

\begin{document}


\begin{frontmatter}
	\title{From Noisy Labels to Intrinsic Structure: A Geometric-Structural Dual-Guided Framework for Noise-Robust Medical Image Segmentation}%
	
	\author[1,2]{Tao Wang\fnref{cofirst}}
	\author[2]{Zhenxuan Zhang\fnref{cofirst}}
	\author[1]{Yuanbo Zhou}
	\author[1]{Xinlin Zhang}
	\author[1]{Yuanbin Chen}
	\author[6]{Tao Tan}
        \author[2,3,4,5]{Guang Yang\fnref{colast}}
        \author[1]{Tong Tong\fnref{colast}\corref{cor1}}
        \fntext[cofirst]{These authors contributed equally to this work.}
        \fntext[colast]{Co-last senior authors.}
	\cortext[cor1]{Corresponding author: ttraveltong@gmail.com}
    
	\address[1]{College of physics and information engineering, Fuzhou University, Xueyuan Road No.2, Fuzhou, 350108, China}
	\address[2]{Bioengineering Department and Imperial-X, Imperial College London, London W12 7SL, UK}
	\address[3]{National Heart and Lung Institute, Imperial College London, London SW7 2AZ, UK}
	\address[4]{Cardiovascular Research Centre, Royal Brompton Hospital, London SW3 6NP, UK}
	\address[5]{School of Biomedical Engineering \& Imaging Sciences, King's College London, London WC2R 2LS, UK}
	\address[6]{Faculty of Applied Science, Macao Polytechnic University, Macao Special Administrative Region of China}
	
	\begin{abstract}
		The effectiveness of convolutional neural networks in medical image segmentation relies on large-scale, high-quality annotations, which are costly and time-consuming to obtain. Even expert-labeled datasets inevitably contain noise arising from subjectivity and coarse delineations, which disrupt feature learning and adversely impact model performance. To address these challenges, this study proposes a Geometric–Structural Dual-Guided Network (GSD-Net), which integrates geometric and structural cues to improve robustness against noisy annotations. It incorporates a Geometric Distance-Aware module that dynamically adjusts pixel-level weights using geometric features, thereby strengthening supervision in reliable regions while suppressing noise. A Structure-Guided Label Refinement module further refines labels with structural priors, and a Knowledge Transfer module enriches supervision and improves sensitivity to local details. To comprehensively assess its effectiveness, we evaluated GSD-Net on six publicly available datasets: four containing three types of simulated label noise, and two with multi-expert annotations that reflect real-world subjectivity and labeling inconsistencies. Experimental results demonstrate that GSD-Net achieves state-of-the-art performance under noisy annotations, achieving improvements of {1.58}\% on Kvasir, 22.76\% on Shenzhen, 8.87\% on BU\_SUC, and {1.77}\% on BraTS2020 under $S_{R}$ simulated noise. The codes of this study are available at \url{https://github.com/ortonwang/GSD-Net}.
	\end{abstract}
	
	
\end{frontmatter}
\begin{figure*}[t]
	\centering
	\includegraphics[width=1\textwidth]{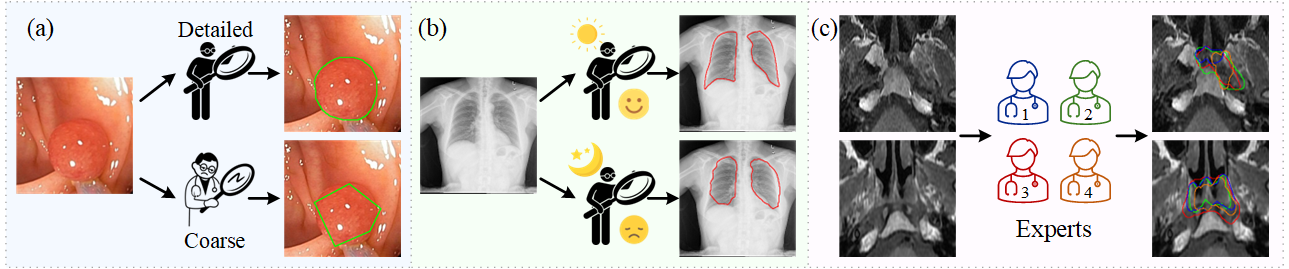}		
	\caption{Illustration of label noise sources: (a) coarse annotations, (b) intra-observer variability (inconsistencies by the same expert under different conditions), and (c) inter-observer variability (annotations from different experts shown in different colors).
    }
	\label{show_question}
\end{figure*}
\section{Introduction}  
Medical image segmentation is a foundational technique in image analysis, which plays a critical role in disease diagnosis and clinical decision-making. 
Convolutional neural networks (CNNs) have been widely adopted for their strong feature extraction capabilities, with U-Net {{\citep{unet}}} and its variants becoming the leading architectures in medical image segmentation {{\citep{liu2024vsmtrans,kuang2025lw,3D_unet}}}. However, the performance of these methods relies heavily on the availability of high-quality, precisely annotated pixel-level labels {{\citep{shen2023survey}}}. In practice, challenges such as limited image quality, indistinct object boundaries, and low contrast between foreground and background {{\citep{zhang2021weakly}}} make precise delineation highly challenging. For instance, Computed Tomography (CT) images often suffer from motion artifact {{\citep{ko2021rigid}}}, Magnetic Resonance Imaging (MRI) is prone to Rician noise and intensity inhomogeneity {{\citep{mehrnia2024novel}}}, while ultrasound commonly exhibits speckle noise and acoustic shadowing {{\citep{singla2022speckle}}}. These modality-specific degradations further obscure anatomical boundaries and complicate accurate segmentation.
As a result, reference annotations are typically provided by experienced clinicians through manual delineation. 

However, this process is susceptible to label noise due to imprecise and biased annotations {{\citep{ma2023adjustable}}}. 
First, pixel-level labeling is costly and time-consuming. To save effort, annotators may produce coarse delineations {{\citep{yu2020robustness}}}. 
As shown in Fig. \ref{show_question} (a), a detailed annotation required 32 seconds, whereas a coarse one took only 5 seconds. Second, the inherent properties of medical images, including low contrast and indistinct boundaries, further increase the risk of boundary-localized errors. 
Third, annotation variability arises both within and between annotators. Intra-observer variability occurs when the same annotator produces inconsistent results due to timing, cognitive state, or emotional condition (Fig. \ref{show_question} (b)). Inter-observer variability occurs when different experts provide inconsistent annotations due to subjectivity, fatigue, or inconsistent region definitions (Fig. \ref{show_question} (c)) {{\citep{wu2024diversified,schmidt2023probabilistic}}}.
In large-scale projects, these variabilities accumulate and generate noisy labels which mislead the training process. Unlike classification noise, which is global and categorical, segmentation noise is spatially localized and structure-dependent. It often occurs near ambiguous boundaries or in anatomically complex regions {{\citep{fang2023reliable}}}. Pixel-level noise further distorts spatial features and impairs the model’s ability to capture object morphology and boundaries. As a result, both robustness and generalizability are diminished {{\citep{yi2021learning}}}. Therefore, mitigating label noise and improving robustness under noisy supervision remain critical challenges in medical image segmentation.

Extensive research has been conducted to address the performance degradation of deep learning models caused by noisy labels. Existing methods can be generally categorized into three categories: 1) Noise-robust loss functions, which modify the loss formulation to reduce the impact of mislabeled samples {{\citep{barron2019general, zhang2018generalized}}}. These approaches are simple to integrate and can effectively down-weight noisy data, but they rely heavily on loss statistics, which may misclassify hard examples as noise and fail to capture structural dependencies. 2) Sample selection, which identifies and prioritizes reliably annotated data {{\citep{fang2023reliable, han2018co}}}. Such strategies mitigate the influence of noisy supervision by focusing on low-loss or high-confidence regions, yet they often discard ambiguous boundary pixels and underutilize supervision from unselected samples. and 3) Label correction, which improves noisy annotations using model predictions {{\citep{yang2022estimating}}}. This paradigm can progressively refine noisy labels and recover useful supervision, but it is sensitive to early model errors and, without explicit structural constraints, may generate artifacts or discontinuous boundaries.

While these approaches have demonstrated strong performance in image classification, directly transferring them to segmentation is non-trivial. 
{Label noise is typically independent in classification but becomes spatially structured in segmentation, particularly in ambiguous regions {\citep{yao2023learning,fang2023reliable}}. 
Due to the lack of explicit modeling of such structured noise and the uneven distribution of reliable supervision across objects, informative pixels may be over-penalized or mistakenly discarded {\citep{HUANG2025106850}}. 
As a result, important structural information is lost, making it challenging for loss-based or sample-selection strategies to preserve reliable information in pixel-level tasks.
}
Moreover, label correction methods are prone to error accumulation and may yield anatomically inconsistent results, underscoring the need for segmentation-specific, structure-aware solutions.
These limitations prevent them from providing reliable supervision under diverse and clinically realistic noise conditions. Despite recent advances, several critical issues remain:
1) Limited noisy-pixel detection. Many approaches rely solely on loss-based criteria (e.g., the small-loss strategy, {which assumes that pixels with smaller losses are more likely correct, and uses only these pixels for model updates to mitigate the effect of label noise {\citep{fang2023reliable}}}), which may discard correctly labeled yet ambiguous boundary pixels while retaining mislabeled regions with artificially low loss due to early overfitting {{\citep{shi2024survey}}}.
2) Insufficient anatomical modeling. Current methods typically rely on local smoothing or morphological constraints, failing to explicitly capture anatomical topology and regional coherence. This often results in fragmented boundaries, shape distortions, and spurious contours under severe noise.
3) Lack of cross-sample complementarity. Most strategies operate at the single-sample level and disregard structural information across images, thereby limiting their ability to recover reliable supervision.
Additionally, many studies simulate label noise using simple morphological operations (e.g., erosion, dilation) {{\citep{gonzalez2023robust, li2021superpixel}}}, which is overly simplistic and fails to capture the diverse, irregular annotations common in clinical practice.
Taken together, these observations underscore the urgent need for noise-robust segmentation methods capable of delivering reliable performance under diverse and clinically realistic conditions.

\begin{figure}[!h]
	\centering
	\includegraphics[width=0.49\textwidth]{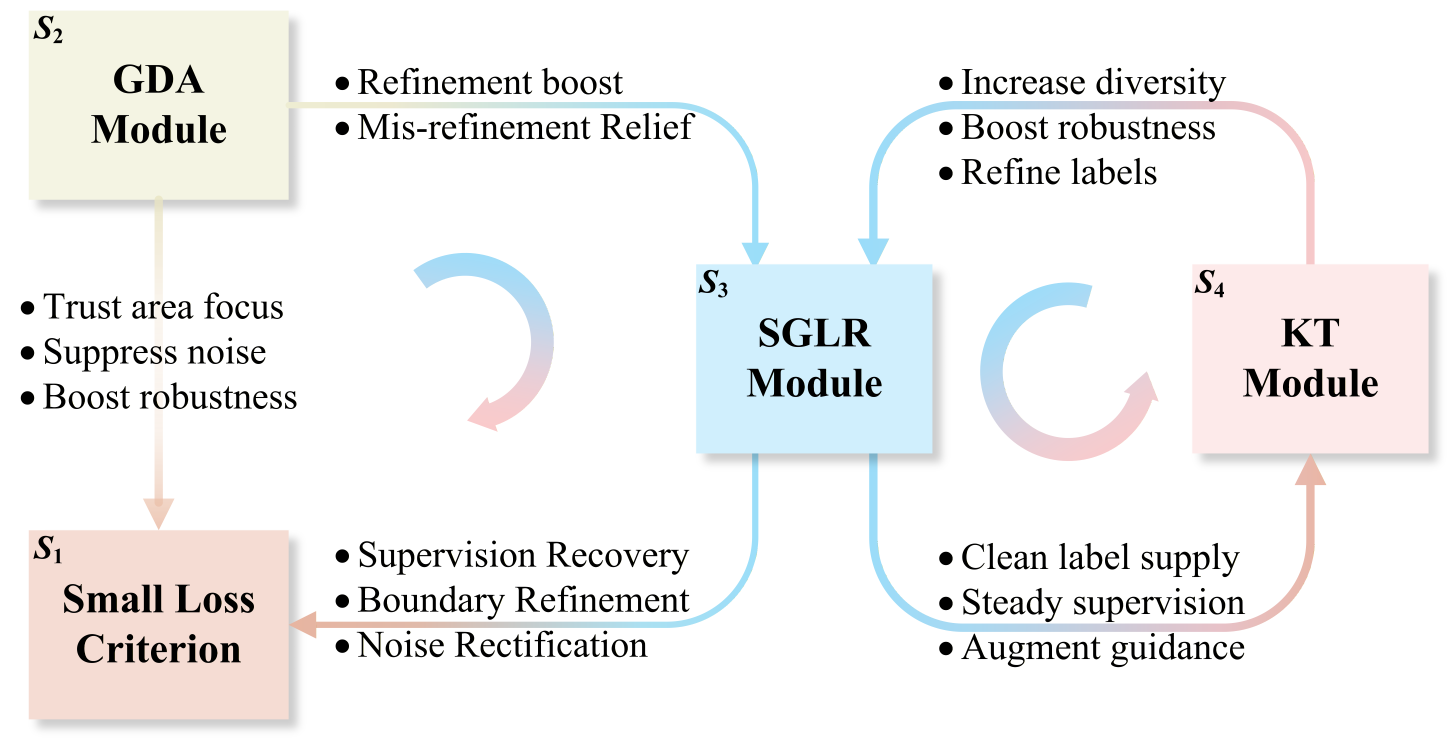}		
	\caption{Illustration of collaborative learning among modules. {Small Loss Criterion (Sec. {\ref{method_jocor}})}; GDA: Geometric Distance-Aware {(Sec. {\ref{Probabilistic_aware}})}; SGLR: Structure-Guided Label Refinement {(Sec. {\ref{dynamic_fusion}})}; KT: Knowledge Transfer {(Sec. {\ref{transfer_section}})}. The texts along the arrows describe the functional effects that one module exerts on another. $S_{\mathrm{1}}$-$S_{\mathrm{4}}$ denote different stages.}
	\label{figure_module_work}
\end{figure}

{
To address the challenges posed by noisy annotations, we propose the Geometric-Structural Dual-Guided Network (GSD-Net), a unified framework that integrates multiple complementary modules in a collaborative and mutually reinforcing manner (Fig.~{\ref{figure_module_work}}). The overall pipeline progressively enhances supervision reliability through small-loss-based region selection, Geometric Distance-Aware (GDA) , Structure-Guided Label Refinement (SGLR), and Knowledge Transfer (KT), forming a synergistic learning process.
Specifically, GSD-Net first employs a dual-model co-regularization strategy combined with the small-loss criterion to identify reliable regions. By selecting samples with relatively small training losses, this step filters out a large portion of noisy annotations and provides an initial set of trustworthy supervisory signals for subsequent refinement.
Building upon these reliable regions, the GDA module introduces geometric priors to reweight the training loss, encouraging the model to focus on reliable areas while suppressing noisy regions. This geometry-aware weighting further guides the small-loss mechanism toward more reliable region selection and stabilizes the subsequent label refinement process.
Subsequently, the SGLR module dynamically fuses the predictions of two networks under superpixel-based structural constraints to correct noisy labels and recover supervisory signals that skipped by the small-loss strategy. Meanwhile, guidance from the GDA module helps alleviate potential mis-refinement, preventing the model from overfitting to unreliable pseudo-labels.
Furthermore, the KT module transfers local structural information across randomly paired images, enhancing model robustness and data diversity. The improved robustness enables the SGLR module to produce more reliable label refinements. In turn, the refined labels generated by SGLR further benefit the KT process, forming a collaborative loop that reinforces both label refinement and knowledge transfer.
Through this progressive and synergistic design, the modules form a positive feedback loop that enhances training robustness. As a result, GSD-Net achieves reliable segmentation performance under clinically realistic noisy annotation conditions.}

The main contributions of this study are summarized as follows:
\begin{itemize}
    \item We propose GSD-Net, a unified framework for noise-robust medical image segmentation that holistically integrates reliability, structural guidance, and cross-sample diversity into a forward-collaborative design.  
    \item This framework enhances robustness through a synergistic mechanism that consolidates reliable supervision with geometric priors, refines noisy regions via structural constraints, and enriches supervision by transferring diverse local patterns across images, with these processes working in concert within a unified framework.
    \item Extensive experiments on six public datasets, including simulated noise and real annotation noise caused by inter-expert variability, demonstrate that our method achieves robust performance and effectively mitigates inter-expert inconsistencies. 
\end{itemize}

\section{Related Work}
Numerous studies have demonstrated that label noise can severely degrade the generalization of deep neural networks, with high-parameter models such as CNNs and Transformers being especially susceptible to overfitting {{\citep{zhang2020robust, xu2022anti, northcutt2021confident}}}. To address this, researchers have developed various noise-robust learning approaches aimed at sustaining high segmentation performance despite noisy annotations. Existing work primarily focuses on the following aspects:
\subsection{Noise-Robust Loss Functions}
In noisy-label learning, loss function design is critical for ensuring model robustness and generalization. Standard cross-entropy (CE) loss assigns equal weight to all samples, which often causes overfitting to noisy labels \citep{zhang2020robust, northcutt2021confident}. To address this, several noise-robust loss functions have been proposed. The Generalized Cross-Entropy (GCE) loss {{\citep{zhang2018generalized}}} uses a tunable parameter to interpolate between mean absolute error and CE loss, balancing robustness and convergence. The Symmetric Cross-Entropy loss {{\citep{wang2019symmetric}}} combines CE with reverse cross-entropy to reduce degradation under both symmetric and asymmetric noise. The Dynamics-Aware Loss {{\citep{li2023dynamics}}} adapts to the learning dynamics of deep networks by emphasizing easy examples in early training, progressively enhancing robustness, and incorporating a bootstrapping term. The multi-class unhinged loss and its smooth variants, SGCE and $\alpha$-MAE, stem from a decomposition theory of multi-class losses, enabling smooth transitions between unhinged loss and MAE for dynamic robustness control {{\citep{paquinsymmetrization}}}. These approaches have shown strong performance in classification tasks, offering valuable insights for medical image segmentation. In segmentation, {{\cite{gonzalez2023robust}}} proposed T-Loss, a robust loss function derived from the Student-t distribution, which adaptively controls sensitivity to label noise and outliers via a learnable parameter. These losses have demonstrated effectiveness in classification tasks, where label noise is less spatially correlated. However, segmentation requires dense, pixel-level predictions and is more sensitive to structured and boundary-localized noise, limiting its effectiveness in this setting.

\subsection{Reliable Sample Selection}
Small-loss sample selection is a widely adopted strategy for noise-robust learning. During early training, models tend to fit correctly labeled “clean” samples first, while noisy samples typically yield higher loss values {{\citep{arpit2017closer}}}. Building on this idea, {{\cite{han2018co}}} introduced the Co-Teaching framework, where two networks are jointly trained and exchange small-loss samples for mutual supervision. Building on Co-Teaching, {{\cite{wei2020combating}}} proposed JoCoR, which adds co-regularization to maintain prediction consistency alongside low-loss selection. 

{{\cite{zhang2020robust}}} extended Co-teaching by training three networks simultaneously, with each pair collaboratively selecting reliable samples to guide the third. {{\cite{fang2023reliable}}} further proposed a collaborative learning framework, where two models clean and distill reliable knowledge from each other using consistency-based regularization. Despite their effectiveness and structural simplicity, these methods rely on loss values as indicators for label correctness, which may lead to the unintended removal of "high-loss clean samples" near ambiguous boundaries or complex structures. This not only restricts the model’s ability to learn fine-grained details but also ignores potentially valuable image information present in noisy regions.

\subsection{Label Correction}
Label correction and pseudo-label generation aim to improve supervision by refining suspected label errors during training. Unlike small-loss strategies that focus exclusively on assumed clean samples, these methods seek to exploit informative content from noisy regions, typically by leveraging high-confidence model predictions. For example, {{\cite{xiao2022promix}}} proposed ProMix, which employs matched high-confidence selection to progressively expand the clean sample set. 

{{\cite{qiu2023hierarchical}}}  A multimodal self-training framework to address label inconsistencies between Whole Slide Images and their patches. {{\cite{liu2022adaptive}}} proposed ADELE, which exploits early-learning dynamics to adaptively correct class-wise noise while enforcing multi-scale consistency. {{\cite{shi2021distilling}}} introduced adaptive thresholding with prototype-guided correction for heavily corrupted subsets, and {{\cite{jin2022deep}}} proposed pixel-level correction through noisy pixel estimation. Nevertheless, the predominant reliance on model-predicted confidence makes these methods vulnerable to biased predictions under severe noise, potentially leading to iterative reinforcement of incorrect labels. To address this limitation, our approach integrates confidence estimation with superpixel-based structural priors, thereby constraining correction within structurally coherent regions and reducing the risk of error propagation.

\begin{figure*}[!h]
	\centering
	\includegraphics[width=0.99\textwidth]{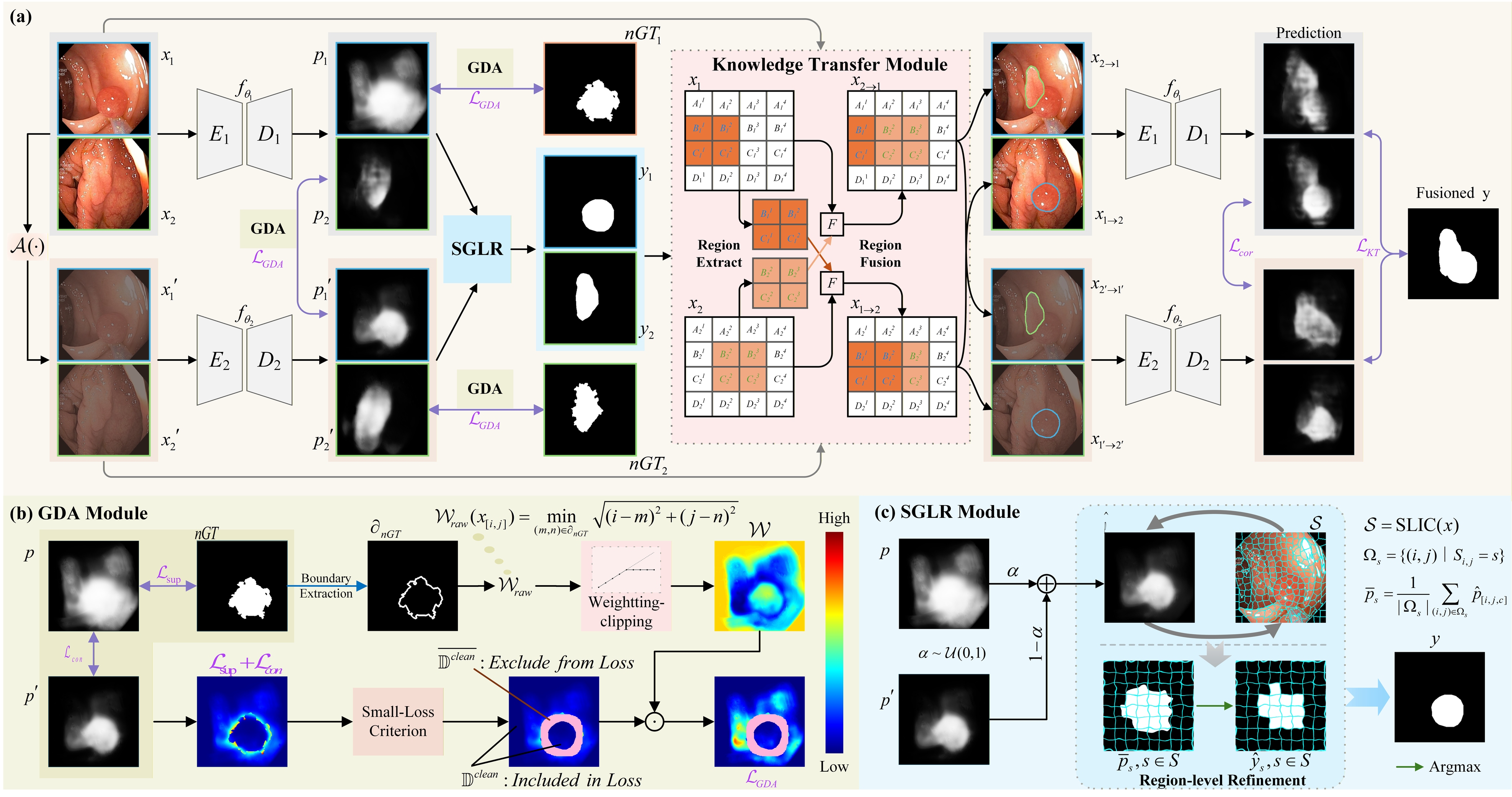}		
	\caption{Schematic diagram of the proposed GSD-Net framework: (a) overall workflow, (b) Geometric Distance-Aware module, and (c) Structure-Guided Label Refinement module. In subfigure (b), $\mathbb{D}^{clean}$ represents the reliable labeled regions selected by the small-loss criterion, while $\overline{\mathbb{D}^{clean}}$ represents the complementary set.}
	\label{figure_framework}
\end{figure*}

\section{Methods}
This study aims to address the challenge of inaccurate pixel-level annotations in medical image segmentation. To facilitate understanding, we first present an overview of the proposed framework, followed by detailed descriptions of its components. Key notations are summarized in Table \ref{tab:notations} for clarity.
\begin{table}[t]
	\centering
	\caption{Descriptions of Key Notations}
	\label{tab:notations}
	\begin{tabular}{cl}
		\toprule
		\textbf{Notations} & \textbf{Descriptions} \\
		\midrule
		$x$, $nGT$ & Input image and its noisy annotation \\
		$\mathcal{A}(\cdot)$ & Weak augmentation operation \\
		$x'$ & Augmented image: $x' = \mathcal{A}(x)$ \\
		$f_{\theta_i}(\cdot)$ & Model with parameters $\theta_i$ \\
		$p_i$ & Prediction from model $i$: $p_i = f_{\theta_i}(x)$ \\
		$\mathbb{D}$ & Pixel coordinate set of $x$ \\
		$\mathbb{D}^{clean}$ & Selected small-loss subset \\
            $\oplus$ & Element-wise addition \\
		$\odot$ & Element-wise multiplication \\
		$\mathcal{S}$ & Generated superpixel map: $\mathcal{S} = \mathrm{SLIC}(x)$ \\
		${y}$ & Generated pseudo-label: ${y} = \mathrm{SGLR}(x)$ \\
		$\mathcal{W}$ & Weight map generated by the GDA module \\
		$\mathcal{KL}$ & Symmetric Kullback$-$Leibler divergence \\
		$x_{1 \to 2}$ & Knowledge transferred from $x_{1}$ to $x_{2}$ \\
		$x_{2 \to 1}$ & Knowledge transferred from $x_{2}$ to $x_{1}$ \\
		\bottomrule
	\end{tabular}
\end{table}

\subsection{Overview}
The overall architecture and pseudo-code of GSD-Net are presented in Fig.~\ref{figure_framework} and Algorithm~\ref{algorithmix1}. Given an input pair ($x_{1}$, $x_{2}$), weak augmentations produce ($x_{1}'$, $x_{2}'$) with predictions
obtained as $p_{1}=f_{\theta_{1}}(x_{1})$, $p_{2}=f_{\theta_{2}}(x_{2})$, $p_{1}'=f_{\theta_{1}}(x_{1}')$, and $p_{2}'=f_{\theta_{2}}(x_{2}')$. A robust co-regularization mechanism is subsequently employed to optimize the model parameters. We first apply a small-loss strategy to identify reliably annotated regions in the simulated noisy ground truth ($nGT$). Then the GDA module suppresses potential noise and reinforces supervision in trustworthy areas. Next, a Structure-Guided Label Refinement module leverages superpixel-based spatial priors to dynamically weight predictions from two networks, thereby refine $nGT$. Finally, the Knowledge Transfer module enriches data diversity and improves the model’s sensitivity to local details and structural variations across samples.

\subsection{Preliminary research}
\label{method_jocor}
JoCoR {{\citep{wei2020combating}}} employs the co-regularization framework to clean up the training data. This method primarily incorporates the small-loss strategy with a contrastive loss and serves as a baseline in our study. The implementation begins by evaluating the loss between predictions and $nGT$, as formulated:
\begin{align}
	\mathcal{L}_{CE}(p_{[i, j]},y_{[i, j]}) = - \sum_{c=1}^{C} y_{[i, j,c]} \log(p_{[i, j,c]}),
    \label{loss_for_ce}
\end{align}
where $y_{[i, j,c]}$ indicates whether the pixel at position $[i, j]$ belongs to the $c$-th class, and $C$ denotes the total number of classes. Similarly, $p_{[i, j,c]}$ represents the probabilities of the pixel at position $(i, j)$ belonging to the $c$-th class. Therefore, the supervised loss is defined as:
\begin{align}
	\mathcal{L}_{sup}(x,nGT) = \mathcal{L}_{{CE}}\left\{f_{\theta_{1}}(x_{[i, j]}),nGT_{[i, j]}\right\}+  \notag \\
	\mathcal{L}_{{CE}}\left\{f_{\theta_{2}}(x'_{[i, j]}),nGT_{[i, j]}\right\},  
	\label{loss_sup}
\end{align}
According to the agreement maximization principle {{\citep{sindhwani2005co}}}, models tend to agree more on correctly labeled regions than on mislabeled ones. JoCoR leverages a contrastive term for co-regularization, where agreement is quantified using the symmetric Kullback-Leibler ($\mathcal{KL}$) divergence, defined as:
\begin{align}
	\mathcal{KL}(P_{[i, j]}||Q_{[i, j]}) = \sum_{c=1}^{C} P_{[i, j,c]} \log \left( \frac{P_{[i, j,c]}}{Q_{[i, j,c]}} \right) ,
	\label{equation_kl}
\end{align}
Therefore, the agreement between the two networks is defined as:
\begin{align}
	\mathcal{L}_{con}(x_{[i, j]}) = \mathcal{KL}\left\{f_{\theta_{1}}(x_{[i, j]}) \parallel f_{\theta_{2}}(x_{[i, j]}')\right\} + \notag \\
	\mathcal{KL}\left\{f_{\theta_{2}}(x_{[i, j]}')\parallel f_{\theta_{1}}(x_{[i, j]})\right\},
	\label{loss_con}
\end{align}
Next, JoCoR selects "clean" regions $\mathbb{D}^{clean}$ using the "small-loss" criterion:
\begin{align}
	\displaystyle
	\mathbb{D}^{clean} = \arg\min_{\mathbb{D}' : |\mathbb{D}'| \geq \mathcal{R}(e)|\mathbb{D}|} \, \sum_{(i,j)\in \mathbb{D}} \mathcal{L}_{sup}(x_{[i, j]}) \oplus \mathcal{L}_{con}(x_{[i, j]}),
	\label{D_clean}
\end{align}
The retention rate is defined as $\mathcal{R}(e)=1-\min(\frac{e}{10}\tau,\tau)$, where $e$ denotes the current epoch and $\tau$ is a constant. At the beginning, $\mathcal{R}(e)$ is close to 1 and it decreases toward $1-\tau$ as training progresses, gradually reducing the selected regions to mitigate overfitting to $nGT$ {{\citep{han2018co}}}. Therefore, the following loss for JoCoR is defined as:
\begin{align}
	\displaystyle
	\mathcal{L}_{JoCoR}(x_{[i,j]},nGT) = \frac{\sum_{(i,j) \in \mathbb{D}^{clean}} \mathcal{L}_{sup}(x, nGT) \oplus \mathcal{L}_{con}(x_{[i, j]})}{|\mathbb{D}^{clean}|} , 
	\label{loss_jocor}
\end{align}
\begin{algorithm}[t]
	\caption{A Geometric-Structural Dual-Guided Framework for Noise-Robust Medical Image Segmentation (Training)}
	\begin{algorithmic}[1]
		\State \textbf{Input:} Training set $\{(x_k, nGT_k)\}_{k=1}^{n}$
		\State \textbf{Initialize:} Two networks $f_{\theta_1}(\cdot)$ and $f_{\theta_2}(\cdot)$
		\For{epoch = 1 to MaxEpoch}
		\State $x'\gets \mathcal{A}(x)$  
		\State  $p \gets f_{\theta_1}(x)$, \quad $p' \gets f_{\theta_2}(x')$
		\State Compute $\mathcal{L}_{sup}$ (Eq.~\eqref{loss_sup}) and  $\mathcal{L}_{con}$ (Eq.~\eqref{loss_con})
		\State $\mathbb{D}^{clean} \gets$ Small-loss Criterion ($\mathcal{L}_{sup}\oplus\mathcal{L}_{con}$)(Eq.~\eqref{D_clean})
		\State  $\mathcal{W} \gets \text{GDA}(nGT, \mathrm{epoch})$ (Eq.~\eqref{eq:wd_weight_clipped}) 
		\State \textcolor[HTML]{0000A0}{// Reweighting Loss $\mathcal{L}_{GDA}$:}
		\State  $\displaystyle \mathcal{L}_{GDA} \gets \frac{\sum_{(i,j) \in \mathbb{D}^{clean}} \left(\mathcal{L}_{sup} + \mathcal{L}_{con}\right) \cdot \mathcal{W}}{|\mathbb{D}^{clean}|} $
		\State \textcolor[HTML]{0000A0}{// Structure-Guided Label Refinement:}
		\State $\mathcal{S} \gets \text{SLIC}(x)$
		\State $y \gets \mathrm{SGLR}(p, p',\mathcal{S},\mathbb{D}^{clean})$ (Sec. \ref{dynamic_fusion})
		\State \textcolor[HTML]{0000A0}{// Knowledge Transfer:}
		\State Sample $x_1, x_2 \sim \{x_k\}_{k=1}^{n}$
		\State $x_{2\to 1}, x_{1\to 2}, y_{2\to 1}, y_{1\to 2} \gets \text{KT}(x_1,x_2,y_1,y_2)$ (Sec. \ref{transfer_section})
		\State  $p_{2\to 1} \gets f_{\theta_1}(x_{2\to 1})$, \quad $p_{2\to 1}' \gets f_{\theta_1}(x_{2\to 1}')$
		\State  $p_{1\to 2} \gets f_{\theta_2}(x_{1\to 2})$, \quad $p_{1\to 2}' \gets f_{\theta_2}(x_{1\to 2}')$
		\State Compute $\mathcal{L}_{KT}$ (Eq.~\eqref{loss_kt}) and $\mathcal{L}_{cor}$ (Eq.~\eqref{L_COR})
		\State \textcolor[HTML]{0000A0}{// Total Loss Function:}
		\State $\mathcal{L}_{total} \gets \mathcal{L}_{GDA} + \mathcal{L}_{KT} + \mathcal{L}_{cor}$
		\State Update parameters $\theta_1$, $\theta_2$ using $\mathcal{L}_{total}$
		\EndFor
		\State \textbf{Return:} Trained models $f_{\theta_1}(\cdot)$, $f_{\theta_2}(\cdot)$
	\end{algorithmic}
	\label{algorithmix1}
\end{algorithm}
\subsection{Geometric Distance-Aware Module} 
\label{Probabilistic_aware}
\begin{figure}[t]
	\centering
	\includegraphics[width=0.48\textwidth]{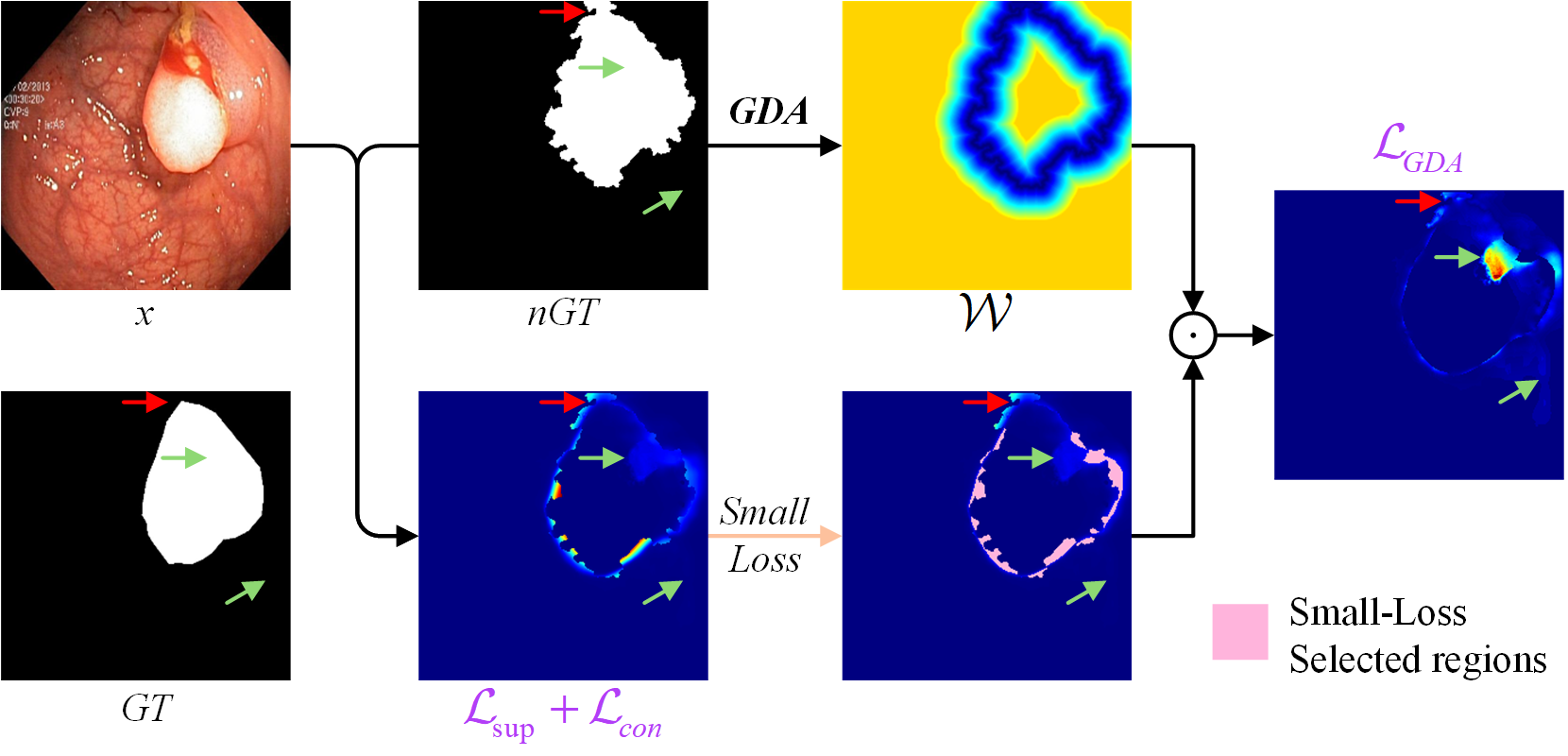}		
	\caption{The schematic diagram of the Geometric Distance-Aware Module.}
	\label{figure_LGDA}
\end{figure}

Although the small-loss criterion is widely adopted for retaining clean regions, it may mistakenly regard high-confidence noisy regions as clean and overlook mislabeled pixels near decision boundaries. To reduce the impact of misidentification, we propose the GDA Module, as illustrated in Fig. \ref{figure_framework} (a) and Fig. \ref{figure_LGDA}, which dynamically adjusts loss weights during training. 
Since annotation errors frequently occur near lesion boundaries {{\citep{lee2020structure}}}, we define a geometric-aware weight map $\mathcal{W}$ that assigns greater weights to pixels farther from the boundaries, which are less likely to be mislabeled. 

{Specifically, for $nGT$, the boundary set $\partial_{nGT}$ is defined as the collection of pixels that exhibit label discontinuity within their immediate neighborhood, formulated as:}
\begin{align}
	\partial_{nGT} = \left\{ (i,j) \in \mathbb{D} \;\middle|\; \exists\, (u,v) \in \mathcal{N}(i,j),\; nGT(u,v) \ne nGT(i,j) \right\},
	\label{eq:boundary}
\end{align}
where $\mathbb{D}$ denotes the set of all pixel coordinates in the image $x$. $\mathcal{N}(i,j)$ represents the {8-connected} neighborhood of pixel $(i,j)$. 
{The condition $nGT(u,v) \ne nGT(i,j)$ ensures that $\partial_{nGT}$ strictly captures the interface coordinates between disjoint semantic classes (e.g., foreground and background).}

{To incorporate geometric awareness, we compute a distance-based weight map $\mathcal{W}_{raw}$ using the Euclidean Distance Transform relative to $\partial_{nGT}$, as formulated:}
\begin{align}
        \displaystyle
	\mathcal{W}_{raw}(x_{[i, j]}) = \min_{(m,n) \in \partial_{nGT}} \sqrt{(i - m)^2 + (j - n)^2}.
	\label{eq:distance_weight}
\end{align}
{
This formulation assigns a scalar to each pixel $(i, j)$ corresponding to its shortest geometric distance to the nearest boundary point $(m, n)$ defined in equation {\eqref{eq:boundary}}. While $\mathcal{W}_{raw}$ provides a global spatial prior, directly utilizing these raw distances may lead the model to overemphasize distant background regions, potentially causing gradient instability or suboptimal convergence on fine structures.}
To mitigate this, we transform $\mathcal{W}_{raw}$ into a dynamic, importance-aware weight map $\mathcal{W}$ through a weight-clipping and temporal decay strategy:
\begin{align}
	\mathcal{W}(x_{[i, j]}; e) =
	\max \bigg<\min\left\{ \mathcal{W}_{raw}(x_{[i, j]}),\; T \right\} - \frac{e}{E}\times T,\; 1\, \bigg> ,
	\label{eq:wd_weight_clipped}
\end{align}
where $T$ represents the threshold for $\mathcal{W}_{raw}$, $e$ denotes the current training epoch, and $E$ denotes the maximum training epochs. 
The refined weight map $\mathcal{W}$ is then integrated into the supervision loss to form the Geometric Distance-Aware Loss $\mathcal{L}_{GDA}$:
\begin{align}
	\mathcal{L}_{GDA} = \frac{1}{|\mathbb{D}^{clean}|}  \sum_{(i, j) \in \mathbb{D}^{clean}} \left\{ \mathcal{L}_{sup}(x_{[i, j]}) \oplus \mathcal{L}_{con}(x_{[i, j]}) \right\} \odot 	\mathcal{W}.
	\label{L_GDA}
\end{align}
The module enhances supervision in reliably labeled regions (green arrows in Fig. \ref{figure_LGDA}) while attenuating the effect of noisy regions (red arrows in Fig. \ref{figure_LGDA}). This is particularly beneficial in the early training stages, where it reduces the influence of label noise and improves both stability and robustness.

\subsection{Structure-Guided Label Refinement Module}
\label{dynamic_fusion}
Although small-loss strategies and GDA modules suppress noisy supervision, they inevitably overlook certain regions whose correction could improve robustness. To address this, we propose the SGLR Module (Fig. \ref{figure_framework} (c)), which integrates predictions from two networks to enhance pseudo-label accuracy and diversity. By enhancing pseudo-label diversity and preserving model independence, the SGLR module mitigates performance degradation from over-enforcing prediction consistency (Eq.~\eqref{equation_kl}). Specifically, a random coefficient $\alpha \sim \mathcal{U}(0,1)$ is used to linearly combine the softmax outputs of the two segmentation branches ($f_{\theta_{1}}$, $f_{\theta_{2}}$), formulated as:
\begin{align}
	\hat{p}_{[i,j,c]} = \bigg<\alpha \cdot \sigma\left\{f_{\theta_{1}}(x_{[i, j]})\right\}\bigg> \oplus \bigg<(1-\alpha)\cdot \sigma\left\{f_{\theta_{2}}(x'_{[i, j]})\right\}\bigg>,
\end{align}
where $\sigma\{\cdot\}$ denotes the softmax operation. To improve pseudo-label reliability, we incorporate structural priors extracted using the SLIC superpixel algorithm {{\citep{achanta2012slic}}}, i.e., $\mathcal{S} = \text{SLIC}(x)$. {Distinct from methods that treat superpixels as hard pseudo-labels, we utilize them as auxiliary structural constraints to regularize the learning process. The algorithm is configured with $n_{\mathrm{segments}}=1024$ and $\text{compactness}=10$, balancing boundary adherence with computational efficiency while preserving anatomically coherent shapes.} For each superpixel region, let $\Omega_s = \{(i,j) \mid S_{i,j} = s\}$ denote its pixel set. The confidence of superpixel $s$ is computed as:
\begin{align}
	\overline{p}_s = \frac{1}{|\Omega_s|} \sum_{(i,j) \in \Omega_s} \hat{p}_{[i,j,c]},
\end{align}
and assign pseudo-labels by selecting the most probable class:
\begin{align}
	\hat{y}_{[i,j]} = \arg\max_{c} \hat{p}_{[i,j,c]}, \quad \text{for } (i,j) \in \Omega_s.
\end{align}
The final pseudo-labels used for training are defined as:
\begin{align}
	y = (\mathbb{D}^{clean} \odot nGT) \oplus (\overline{\mathbb{D}^{clean}}\odot \hat{y}),
\end{align}
where $\mathbb{D}^{clean}$ denotes the clean regions as defined by Eq.~\eqref{D_clean}, and $\overline{\mathbb{D}^{clean}}$ represents their complement, i.e., $\overline{\mathbb{D}^{clean}} = 1 - \mathbb{D}^{clean}$.

\subsection{Knowledge Transfer Module} \label{transfer_section}
To enhance diversity and sensitivity to local structures, we propose a Knowledge Transfer (KT) module. {By introducing controlled cross-sample structural perturbations, the module strengthens noise robustness and facilitates noisy label correction in the SGLR module.
Specifically,} KT module extracts local regions from randomly paired images and embeds them into their counterparts,{generating structurally perturbed samples composed of locally realistic regions from different images. This strategy exposes the model to diverse structural contexts, enabling it to capture fine-grained anatomical details and structural variability, and thus learn more robust and generalizable feature representations under noisy supervision.}
Let $\mathcal{M}_1$ and $\mathcal{M}_2$ denote the masks of regions extracted from $x_{1}$ and $x_{2}$, respectively. Each mask is generated by randomly sampling a foreground or background region {according to the pseudo-labels $y$}, with the sampling probability determined by the proportion of the target foreground area in the image.
{The fusion process generates composite samples by transplanting regions from a source image $i$ onto a target image $j$, which is formulated as:
}
\begin{align}
	x_{i \to j} = x_j \odot (1 - \mathcal{M}_i) + x_i \odot \mathcal{M}_i, \quad i,j \in \{1,2\} \quad i \neq j, \notag \\
    y_{i \to j} = y_j \odot (1 - \mathcal{M}_i) + y_i \odot \mathcal{M}_i, \quad i,j \in \{1,2\}, \quad i \neq j,  \notag \\
    \mathcal{W}_{i \to j} = \mathcal{W}_j \odot (1 - \mathcal{M}_i) + \mathcal{W}_i \odot \mathcal{M}_i, \quad i,j \in \{1,2\} \quad i \neq j, 
	\label{equation_bcp}
\end{align}
{where $x_{i \to j}$ denotes the synthesized image with semantic content transferred from $x_i$ to $x_j$. The corresponding pseudo-labels $y_{i \to j}$ and weight maps $\mathcal{W}_{i \to j}$ are generated using the same fusion strategy.}

{The fused images $x_{i \to j}$ are then fed into the networks for supervision. The total supervision loss after the KT module, combining the Cross-Entropy loss ($\mathcal{L}_{CE}$, as previously formulated in Eq. {\eqref{loss_for_ce}}) and the Dice loss ($\mathcal{L}_{DC}$). Specifically, $\mathcal{L}_{DC}$ is employed to ensure structural consistency and mitigate class imbalance, defined as:}

\begin{align} 
\displaystyle
\mathcal{L}_{DC}(p, y) = 1 - \frac{2 \cdot \sum_{i,j} \sum_{c=1}^{C} p_{[i, j, c]} \cdot y_{[i, j, c]} + \epsilon}{\sum_{i,j} \sum_{c=1}^{C} p_{[i, j, c]}^2 + \sum_{i,j} \sum_{c=1}^{C} y_{[i, j, c]}^2 + \epsilon}, 
\end{align}
{where $p_{[i, j, c]}$ and $y_{[i, j, c]}$ represent the predicted probability and the ground-truth label for class $c$ at coordinate $(i, j)$, respectively, and $\epsilon$ is a smoothing term to ensure numerical stability. By integrating these components, the total supervision loss for the Knowledge Transfer module is formulated as:}
\begin{align}
	\mathcal{L}_{KT} = \ 
	& \sum_{(x, y, \mathcal{W}) \in \mathcal{P}}
	\big\{
	\mathcal{L}_{CE}\left[f_{\theta_1}(x), y\right] \odot \mathcal{W} +
	\mathcal{L}_{DC}\left[f_{\theta_1}(x), y\right]
	\big\} \notag \\
	+ \ 
	& \sum_{(x, y, \mathcal{W}) \in \mathcal{P}'}
	\big\{
	\mathcal{L}_{CE}\left[f_{\theta_2}(x), y\right] \odot \mathcal{W} +
	\mathcal{L}_{DC}\left[f_{\theta_2}(x), y\right]
	\big\},
	\label{loss_kt}
\end{align}
where the pairwise transformation sets are defined as follows:
\begin{align}
	\mathcal{P} &= \left[(x_{1\to2}, y_{1\to2}, \mathcal{W}_{1\to2}),\ 
	(x_{2\to1}, y_{2\to1}, \mathcal{W}_{2\to1})\right], \notag \\
	\mathcal{P}' &= \left[(x_{1'\to2'}, y_{1'\to2'}, \mathcal{W}_{1\to2}),\ 
	(x_{2'\to1'}, y_{2'\to1'}, \mathcal{W}_{2\to1})\right]. \notag
\end{align}
We further incorporate a co-regularization mechanism to encourage consistency between networks, as formulated:
\begin{align}
	\mathcal{L}_{cor} =  \mathcal{KL}\left[f_{\theta_{1}}(x_{2\to 1})\parallel f_{\theta_{2}}(x_{2'\to 1'})\right]\cdot \mathcal{W}_{2\to 1} + \notag \\ 
	\mathcal{KL}\left[f_{\theta_{1}}(x_{1\to 2})\parallel f_{\theta_{2}}(x_{1'\to 2'})\right] \cdot \mathcal{W}_{1\to 2} ,
	\label{L_COR}
\end{align}
this term enforces agreement on transferred samples while accounting for geometric-aware weights.
\subsection{Total Loss Function}
For end-to-end optimization, we integrate all module-specific losses into a unified objective, defined as:
\begin{align}
	\mathcal{L}_{total} =  \mathcal{L}_{GDA}+ \mathcal{L}_{KT} + \mathcal{L}_{cor}, 
\end{align}
where $\mathcal{L}_{GDA}$, $\mathcal{L}_{KT}$, and $\mathcal{L}_{cor}$ are defined in Eq.~\eqref{L_GDA}, \eqref{loss_kt}, and \eqref{L_COR}, respectively.

\section{Experiments and Results}
\subsection{Datasets}
\begin{figure}[t]
	\centering
	\includegraphics[width=0.49\textwidth]{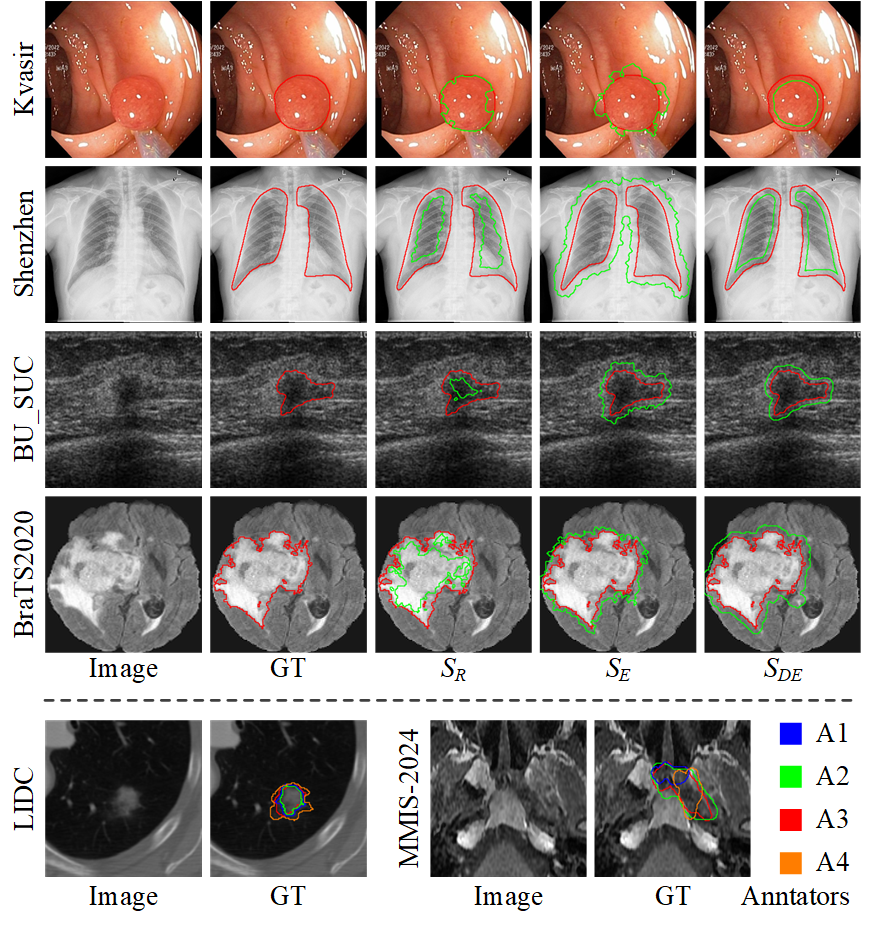}	
	\caption{Visualization of simulated label noise ($S_R$: foreground-reducing, $S_E$: foreground-expanding, $S_{DE}$: simulated via dilation or erosion) and inter-expert variability. In the upper panel, red contours represent ground truth boundaries, and green contours indicate simulated noisy labels.}
	\label{show_simulated_noise}
\end{figure}

We evaluate our method on six publicly available datasets against state-of-the-art approaches. We simulate label noise on Kvasir {{\citep{jha2019kvasir}}}, Shenzhen {{\citep{candemir2013lung, jaeger2013automatic, stirenko2018chest}}}, BU\_SUC {{\citep{iqbal2024memory}}}\footnote{https://www.kaggle.com/datasets/orvile/bus-uc-breast-ultrasound}, and BraTS2020 {{\citep{menze2014multimodal}}} datasets, which all provide precise pixel-level annotations for controlled noise simulation and robustness evaluation.
As widely recognized and representative benchmarks in medical image segmentation, they span diverse imaging modalities such as endoscopy, X-radiation (X-ray), ultrasound (US), and MRI, enabling a comprehensive assessment of the method’s applicability and generalizability across different clinical contexts. To evaluate robustness under real-world noise, we further evaluate on the LIDC {{\citep{armato2011lung}}} and MMIS-2024  \citep{luo2023deep, wu2024diversified, bakas2021multi, cepeda2023rio, suter2022lumiere} datasets, where each image is annotated by multiple independent experts. These datasets capture label noise arising from inter-observer variability (Fig.~\ref{show_simulated_noise}). The details of each dataset are summarized in Table \ref{dataset_introduces}.

\begin{table}[t]
	\centering
        \caption{Summary of the datasets used in our experiments}
        \label{dataset_introduces}
	\resizebox{0.48\textwidth}{!}{
		\begin{threeparttable}
				\begin{tabular}{l|m{1.35cm}<{\centering}|m{4.75cm}|m{2.1cm}|m{0.45cm}<{\centering}m{0.45cm}<{\centering}|c}
					\hline
					Datasets   & Modalities   & Characteristics                                                                           & Target                   & Train& Test & Size  \\ \hline
					Kvasir    & Endoscopy  & \begin{tabular}[c]{@{}l@{}}\textbullet\ Uneven illumination and glare	\\ \textbullet\ Blurry tissue boundaries\end{tabular}   & \begin{tabular}[c]{@{}l@{}}Gastrointestinal \\ polyp\end{tabular}	& 700       & 300        & {$256^{2}$}     \\ \hline
					Shenzhen  & X-Ray      & \begin{tabular}[c]{@{}l@{}}\textbullet\ Foreground$-$background overlap\\ \textbullet\ Low soft-tissue contrast\end{tabular}                     & Lungs area                   & 453       & 113        & {$256^{2}$}     \\ \hline
					
					BU\_SUC     & US &\begin{tabular}[c]{@{}l@{}}\textbullet\ High noise and low contrast\\ \textbullet\ Heterogeneous echogenicity\end{tabular}  & Breast tumor             & 568       & 243        & {$256^{2}$}     \\ \hline
					
					BraTS2020 \tnote{\ding{172}}& MRI        & \begin{tabular}[c]{@{}l@{}}\textbullet\ Heterogeneous tissue \\ \textbullet\ Irregular tumor shapes\\ \end{tabular} & Glioma tumor             & 296       & 73         & {$128^{3}$} \\ \hline
					
					LIDC \tnote{\ding{173}}      & CT         & \begin{tabular}[c]{@{}l@{}}\textbullet\ Diverse nodule sizes and shapes \\ \textbullet\ Adhesion to surrounding vessels\end{tabular}  & \begin{tabular}[c]{@{}l@{}}Pulmonary \\ nodules\end{tabular}        & 1287      & 322        & {$128^{2}$}     \\ \hline
					
					MMIS-2024 \tnote{\ding{174}} & MRI        & \begin{tabular}[c]{@{}l@{}}\textbullet\ Complex anatomical structures \\ \textbullet\ Infiltration into adjacent tissues \end{tabular} & \begin{tabular}[c]{@{}l@{}}Nasopharyngeal \\ carcinoma\end{tabular} & 100      & 20   & {$128^{2}$}     \\ \hline
				\end{tabular}  %
			\begin{tablenotes}
				\footnotesize
				\item[\ding{172}] Experiments use T1, T2, T1ce, and FLAIR sequences, with all tumor subtypes merged into a binary mask.
				\item[\ding{173}] Following \citep{kohl2018probabilistic, wu2024diversified}, 1,609 patches, each annotated by four radiologists were selected.
				\item[\ding{174}] 100 volumes (2405 slices) were used for training and 20 volumes for testing; inference was performed slice-wise and subsequently reconstructed into 3D volumes for evaluation following \citep{wu2024diversified}.
			\end{tablenotes}
	\end{threeparttable} 
}
\end{table}

\subsection{Implementation Details} 
The proposed framework was implemented using PyTorch and evaluated on an NVIDIA RTX 4090 GPU with 24 GB of memory. For 2D segmentation tasks, we used U-Net {{\citep{unet}}} as the backbone with a batch size of 16. For BraTS2020, we adopted 3D U-Net {{\citep{3D_unet}}} with a batch size of 4. All models were trained for 100 epochs using stochastic gradient descent (SGD) optimizer with a learning rate of 5$\times$10$^{-3}$ and a weight decay of 1$\times$10$^{-5}$. The maximum scaling factor $T$ in the GDA module was configured based on dataset resolution, with a value of 10 for 256$\times$256 images and 5 for 128$\times$128 images. 
{Two networks, $f_{\theta_{1}}$ and $f_{\theta_{2}}$, are trained jointly in a unified framework without shared weights. Each network has independent parameters and gradients. During inference, the final prediction is obtained by fusing the predictions of the two networks on the same test image $x$, which can be formulated as $p = f_{\theta_1}(x) + f_{\theta_2}(x)$.} In the small-loss strategy, the constant $\tau$ for the simulated datasets was set according to the actual noise rate estimated from the ground truth in simulated datasets.
For the LIDC and MMIS-2024 datasets, $\tau$ is determined by the 15\% of the foreground proportion, resulting in 0.003 for LIDC and 0.008 for MMIS-2024.

\subsection{Noise simulation}
Many studies use morphological operations (e.g., dilation or erosion) to generate simulated label noise, which we denote as $S_{DE}$. However, such noise is typically spatially uniform and structurally oversimplified, failing to capture the irregular, boundary-focused characteristics of real annotation errors. To better approximate realistic label noise, we employ a Markov-based boundary perturbation strategy {{\citep{yao2023learning}}}, which randomly distorts ground-truth contours.
Building on this, we introduce two additional noise types beyond the commonly used $S_{DE}$: (1) foreground-reducing noise ($S_R$), simulating under-annotation; and (2) foreground-expanding noise ($S_E$), simulating over-annotation. These noise patterns capture irregular, boundary-focused errors often observed in real annotations and enable a more rigorous evaluation of model robustness under both under- and over-annotation scenarios. {Table {\ref{hyperparameters_simulated}} reports the hyperparameters of simulated noisy labels and their Dice scores with respect to the original ground truth, quantifying the corruption level of each noise type.} Examples of the generated noisy labels are illustrated in the upper panel of Fig. \ref{show_simulated_noise}. 
{For all experiments in this study, model checkpoints from the final 10 training epochs were used for evaluation, and performance is reported as mean $\pm$ standard deviation, following the evaluation protocol in T-Loss {\citep{gonzalez2023robust}}.}

\begin{table}[!h]
\caption{{Hyperparameters of Simulated Noise Types and Corresponding Dice Scores with Ground Truth.}}
	\label{hyperparameters_simulated}
	\resizebox{0.5\textwidth}{!}{
    \begin{threeparttable}
\begin{tabular}{ccccc}
\hline
\multicolumn{5}{c}{Hyperparameters for Simulated Noisy Labels}                                                                                                                                                                \\ \hline
\multicolumn{1}{c|}{\makecell{Noise \\ Setting}} & Kvasir                         & Shenzhen                       & BU\_SUC                        & BraTS2019                      \\ \hline
\multicolumn{1}{c|}{$S_R$}                                                                & $\mathcal{M}$ (200,0.2,0.05)   & $\mathcal{M}$ (200, 0.2, 0.05) & $\mathcal{M}$ (200, 0.2, 0.05) & $\mathcal{M}$ (200, 0.4, 0.05) \\
\multicolumn{1}{c|}{$S_E$}                                                                & $\mathcal{M}$ (200, 0.8, 0.05) & $\mathcal{M}$ (200, 0.8, 0.05) & $\mathcal{M}$ (200, 0.8, 0.05) & $\mathcal{M}$ (50, 0.6, 0.05)  \\
\multicolumn{1}{c|}{$S_{DE}$}                                                             & $\mathcal{K}$ (9 - 11)         & $\mathcal{K}$ (9 - 11)         & $\mathcal{K}$ (9 - 11)         & $\mathcal{K}$ (2 - 4)          \\ \hline
\multicolumn{5}{c}{Dice scores between simulated noise ($nGT$) and ground truth ($GT$)}                                                                                                                                                      \\ \hline
\multicolumn{1}{c|}{$S_R$}                                                                & 84.97                          & 61.33                          & 61.46                          & 86.03                          \\
\multicolumn{1}{c|}{$S_E$}                                                                & 88.61                          & 74.68                          & 74.91                          & 89.52                          \\
\multicolumn{1}{c|}{$S_{DE}$}                                                             & 76.14                          & 74.91                          & 79.34                          & 82.99                          \\ \hline
\end{tabular}
\begin{tablenotes}
\footnotesize
\item[$\bullet$] {{\normalsize $\mathcal{M}\ (T,P,V)$: Markov-based boundary perturbation strategy, with parameters $T$ = step number, $P$ = Bernoulli  preference, $V$ = Bernoulli variance.}}
\item[$\bullet$] {{\normalsize$\mathcal{K}(k_1 - k_2)$:  Morphology-based erosion/dilation kernel, randomly varying between $k_1$ and $k_2$ for each sample.}}
\end{tablenotes}
	\end{threeparttable} 
}
\end{table}

\begin{table*}[t]
    \centering
	\caption{Performance comparison on Kvasir, Shenzhen, and BU\_SUC dataset under three simulated label-noise settings.}
	\label{table_Kvasir_shenzhen_busuc}
	\resizebox{0.99\textwidth}{!}{
		\begin{tabular}{l|ccc|ccc|ccc}
			\hline
			\multirow{2}{*}{Method} & \multicolumn{3}{c|}{Kvasir dataset}                                         & \multicolumn{3}{c|}{Shenzhen dataset}                                       & \multicolumn{3}{c}{BU\_SUC dataset}                                           \\ \cline{2-10}
			& $S_R$                   & $S_E$                   & $S_{DE}$                & $S_R$                   & $S_E$                   & $S_{DE}$                & $S_R$                   & $S_E$                   & $S_{DE}$                \\ \hline
			CE Loss     & 66.86$\pm$4.46          & 70.33$\pm$3.25          & 63.11$\pm$2.54          & 60.36$\pm$1.68          & 73.82$\pm$1.58          & 76.08$\pm$3.01          & 58.14$\pm$2.75          & 73.01$\pm$1.52          & 79.54$\pm$2.99          \\
			GCE Loss    & 66.40$\pm$2.63          & 68.04$\pm$1.67          & 60.67$\pm$2.59          & 60.86$\pm$1.01          & 74.83$\pm$0.53          & 76.04$\pm$3.40          & 57.91$\pm$2.23          & 72.99$\pm$0.83          & 78.78$\pm$1.07          \\
			RCE Loss    & 73.51$\pm$1.58          & 73.68$\pm$1.57          & 66.17$\pm$1.97          & 59.64$\pm$2.78          & 74.67$\pm$0.67          & 78.82$\pm$2.67          & 59.37$\pm$2.12          & 73.54$\pm$0.59          & 79.74$\pm$1.74          \\
			RMD         & 68.34$\pm$2.18          & 71.57$\pm$1.09          & 66.90$\pm$1.75          & 60.20$\pm$0.26          & 74.84$\pm$0.09          & 86.88$\pm$0.32          & 48.20$\pm$1.62          & 76.31$\pm$0.48          & 79.10$\pm$1.26          \\
			ADELE       & 60.97$\pm$14.78         & 67.10$\pm$11.42         & 60.62$\pm$13.04         & 60.26$\pm$3.06          & 72.74$\pm$2.29          & 81.62$\pm$8.71          & 57.12$\pm$9.01          & 71.86$\pm$7.65          & 80.63$\pm$10.25         \\
			CDR         & 67.87$\pm$3.51          & 70.58$\pm$1.65          & 63.51$\pm$1.67          & 64.84$\pm$5.39          & 75.69$\pm$1.79          & 79.79$\pm$2.29          & 59.95$\pm$2.63          & 74.53$\pm$0.92          & 80.71$\pm$2.61          \\
			Co-Teaching & 74.26$\pm$1.71          & 75.57$\pm$1.12          & 74.03$\pm$1.48          & 63.10$\pm$0.81          & 75.99$\pm$0.29          & 91.63$\pm$0.15          & 57.09$\pm$0.94          & 76.92$\pm$0.46          & 85.54$\pm$0.91          \\
			IDMPS       & 77.52$\pm$1.21          & 74.16$\pm$0.81          & 69.47$\pm$0.81          & 61.93$\pm$1.19          & 74.69$\pm$0.51          & 78.81$\pm$1.29          & 60.22$\pm$1.96          & 73.92$\pm$0.62          & 82.83$\pm$0.72          \\
			JoCoR       & 67.98$\pm$3.23          & 71.43$\pm$1.37          & 65.62$\pm$1.94          & 67.46$\pm$0.37          & 73.44$\pm$0.07          & 89.56$\pm$0.04          & 56.42$\pm$0.77          & 73.54$\pm$0.28          & 81.31$\pm$0.16          \\
			SP-Guide    & 69.24$\pm$2.09          & 62.97$\pm$1.65          & 61.74$\pm$1.56          & 70.55$\pm$0.63          & 74.92$\pm$0.24          & 81.87$\pm$0.42          & 71.40$\pm$0.74          & 71.07$\pm$0.47          & 81.32$\pm$0.54          \\
            {MSA}                     & {78.46$\pm$0.56}              & {75.30$\pm$0.38}            & {72.34$\pm$0.08}            & {62.86$\pm$0.61}            & {73.05$\pm$0.56}            & {78.45$\pm$0.48}            & {63.41$\pm$0.91}            & {73.52$\pm$0.57}            & {83.20$\pm$0.23}            \\
            {T-Loss}                                      & {71.19$\pm$1.20}                       & {72.05$\pm$1.01}                       & {62.83$\pm$1.37}                                            & {60.56$\pm$1.76}                      & {74.55$\pm$0.64}                       & {74.42$\pm$2.31}                                            & {58.55$\pm$2.10}                       & {73.55$\pm$0.75}                       & {78.35$\pm$1.58}                       \\
			Ours        & \textbf{80.04$\pm$0.42} & \textbf{79.39$\pm$0.33} & \textbf{79.97$\pm$0.53} & \textbf{93.31$\pm$0.09} & \textbf{89.25$\pm$0.25} & \textbf{94.68$\pm$0.07} & \textbf{80.27$\pm$0.68} & \textbf{84.59$\pm$0.41} & \textbf{89.30$\pm$0.25} \\ \hline
\end{tabular} }
\end{table*}

\subsection{Comparison with State-of-the-Art Methods}
To evaluate the effectiveness of our method, we conducted comparisons with a wide range of advanced approaches. These include standard loss functions such as Cross-Entropy (CE) and noise-robust alternatives, such as Generalized Cross Entropy (GCE) {{\citep{zhang2018generalized}}}, Reverse Cross Entropy (RCE) {{\citep{wang2019symmetric}}}, {and T-Loss {\citep{gonzalez2025robust}}, a heavy-tailed robust loss, also included in our evaluation.} We also consider advanced frameworks such as IDMPS {{\citep{zhao2024ultrasound}}}, ADELE {{\citep{liu2022adaptive}}}, and CDR {{\citep{xia2020robust}}}. The comparison also covered co-training and small-loss-based approaches such as Co-Teaching {{\citep{han2018co}}}, JoCoR {{\citep{wei2020combating}}}, and RMD {{\citep{fang2023reliable}}}. Additionally, we compared with SP-Guide {{\citep{li2021superpixel}}}, which leverages superpixels as structural priors for noisy label refinement. {Furthermore, we compared our framework with the Medical SAM Adapter (MSA) {\citep{wu2025medical}}, representing the latest state-of-the-art in adapting foundation models for fully automatic medical image segmentation.}
For all experiments, the Dice score was used as the evaluation metric.
\begin{figure*}[!h]
	\centering
	\includegraphics[width=0.99\textwidth]{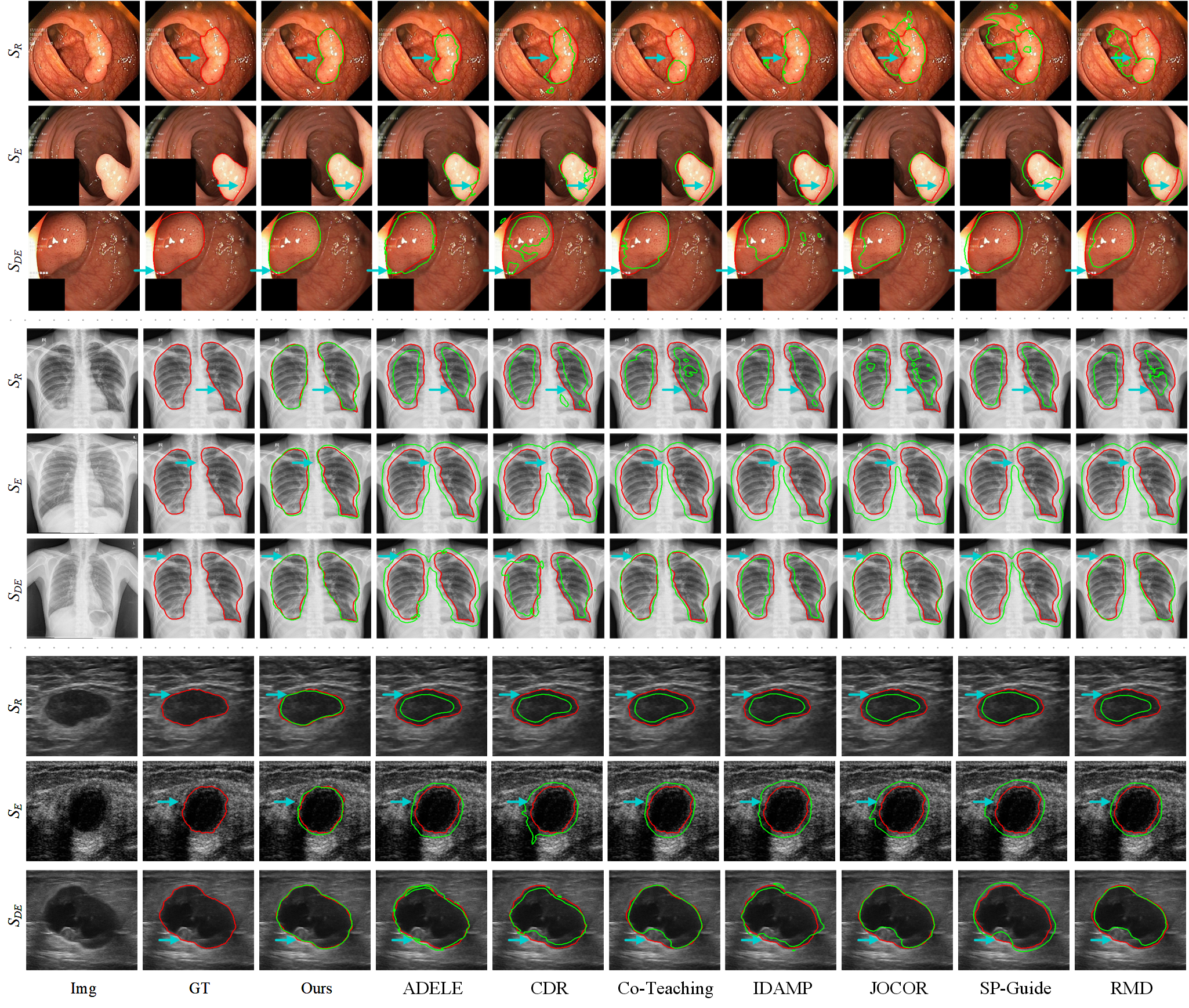}		
	\caption{Qualitative comparison on the Kvasir (upper), Shenzhen (middle) and BU\_SUC (lower) datasets under three simulated label-noise settings. Red contours denote ground-truth boundaries, while green contours represent predicted boundaries.}
	\label{seg_visual_kvasir_shenzhen_busuc}
\end{figure*}

\subsubsection{Results on the Kvasir, Shenzhen, and BU\_SUC datasets}
Table~\ref{table_Kvasir_shenzhen_busuc} summarizes the quantitative results on three datasets under simulated noise conditions ($S_{R}$, $S_{E}$, and $S_{DE}$), with qualitative comparisons provided in Fig.~\ref{seg_visual_kvasir_shenzhen_busuc}. Across all settings, our method consistently outperforms existing approaches, demonstrating robustness across different imaging modalities and adaptability to diverse noise patterns.

On the Kvasir dataset, the improvement is particularly pronounced under the $S_{R}$ noise setting, where compared methods struggle with foreground-reducing noise and exhibit marked Dice score degradation. Our method effectively mitigates this issue and achieves the highest performance, even under the widely used $S_{DE}$ setting, reaching a Dice score of 79.97\%, which surpasses the best competing method at 74.03\%.

On the Shenzhen dataset, which contains larger foreground regions, simulated noise corrupts a greater portion of each image, thereby increasing task difficulty. Our method achieves Dice scores of 93.31\% and 89.25\% under $S_{R}$ and $S_{E}$ noise, respectively, outperforming competing methods that suffer from severe under- and over-segmentation. Even in the commonly adopted $S_{DE}$ setting, where strong baselines such as Co-Teaching reach 91.63\%, our method still achieves the best result with a Dice score of 94.68\%.

On the BU\_SUC dataset, characterized by high speckle noise and small, low-contrast lesions, our method substantially outperforms competing methods, achieving 80.27\% Dice under the challenging $S_{R}$ setting compared with 71.40\% by the best method. While performance improves across all methods under $S_{DE}$ noise, our framework offers superior segmentation by retaining delicate lesion details.

Overall, these results highlight that our framework not only achieves state-of-the-art performance across different datasets but also exhibits strong robustness against diverse and severe noise corruptions.

\subsubsection{Results on the BraTS2020 dataset}
We further evaluated our method on the 3D BraTS2020 brain tumor segmentation dataset. As reported in Table \ref{result_bra}, it consistently outperformed existing approaches across all three simulated noise settings. Specifically, it achieved Dice scores of 81.53\%, 82.68\%, and 83.84\%, respectively, which surpass the best-performing baselines by a notable margin. These results highlight the robustness of our framework in handling complex tumor structures and mitigating annotation noise in 3D brain MRI segmentation task. 
\begin{table}[!h]
	\caption{Performance comparison of GSD-Net and existing methods on the BraTS2020 dataset under three simulated label-noise settings.}
	\label{result_bra}
	\resizebox{0.49\textwidth}{!}{
		\begin{tabular}{l|c|c|c}
			\hline
			Method  & {$S_R$} & {$S_E$} & {$S_{DE}$} \\ \hline
			CE Loss  &	69.16$\pm$1.02 	&	75.61$\pm$0.30	&	67.90$\pm$0.65	\\
			ADELE      & 67.60$\pm$5.07              & 80.82$\pm$1.45              & 71.24$\pm$2.40              \\
			CDR        & 70.00$\pm$3.96              & 79.04$\pm$1.14              & 73.06$\pm$4.21              \\
			Co-Teaching & 76.94$\pm$1.72              & 79.40$\pm$1.55              & 80.81$\pm$1.08              \\
			IDMPS     & 74.01$\pm$4.20              & 78.99$\pm$2.27              & 73.21$\pm$2.05              \\
			JoCoR      & 67.97$\pm$0.44              & 76.36$\pm$0.21              & 76.85$\pm$0.16              \\
            {T-Loss}  & {79.76$\pm$1.73}              & {81.77$\pm$1.53}              & {81.44$\pm$0.98}              \\
			Ours       & \textbf{81.53$\pm$1.44}              & \textbf{82.68$\pm$0.65}              & \textbf{83.84$\pm$0.67}              \\ \hline
	\end{tabular} }
\end{table}

\begin{table*}[]
\caption{{Performance comparison of GSD-Net and existing methods on the LIDC and MMIS-2024 dataset. The A1, A2, A3, and A4 refer to four independent annotators, "Sample" refers to randomly assigning a fixed expert annotation to each image.}}
\label{LIDCMMIS}
\resizebox{\textwidth}{!}{
\begin{tabular}{l|ccc|ccccl|ccccc}
\Xhline{1.5pt}
\multirow{2}{*}{Method}   & \multirow{2}{*}{Ann} & \multirow{2}{*}{\makecell{Image\\Num}} & \multirow{2}{*}{\makecell{Ann\\Cost}} & \multicolumn{5}{c|}{LIDC dataset}                                                  & \multicolumn{5}{c}{MMIS-2024 dataset}                                              \\ \cline{5-14}  
&                      &                                                                                   &                                                                                  & Dice$_{A1}$           & Dice$_{A2}$           & Dice$_{A3}$           & Dice$_{A4}$           & \multicolumn{1}{c|}{Dice$_{mean}$} & Dice$_{A1}$           & Dice$_{A2}$           & Dice$_{A3}$           & Dice$_{A4}$           & Dice$_{mean}$         \\ \hline
\multirow{4}{*}{CE Loss}                        & A1                   & 100$\%$                                                                           & 100$\%$                                                                          & \textbf{88.97$\pm$ 0.27} & 86.21$\pm$0.20          & 86.84$\pm$0.17          & 86.37$\pm$0.27          & \textbf{87.10$\pm$0.23}              & \textbf{85.50$\pm$0.61} & 72.04$\pm$0.35          & 71.64$\pm$0.51          & \textbf{75.47$\pm$0.56} & \textbf{76.16$\pm$0.26} \\
& A2                   & 100$\%$                                                                           & 100$\%$                                                                          & 86.09$\pm$0.18          & \textbf{88.84$\pm$0.21} & 85.99$\pm$0.12          & 85.29$\pm$0.20          & 86.55$\pm$0.18                       & \textbf{75.11$\pm$1.61} & \textbf{74.65$\pm$1.36} & 71.45$\pm$1.36          & 72.27$\pm$1.52          & 73.37$\pm$0.73          \\
& A3                   & 100$\%$                                                                           & 100$\%$                                                                          & 86.73$\pm$0.51          & 85.84$\pm$0.33          & \textbf{88.85$\pm$0.36} & 85.70$\pm$0.44          & 86.78$\pm$0.41                       & 71.18$\pm$2.09          & 71.85$\pm$1.46          & \textbf{74.84$\pm$1.83} & 69.96$\pm$1.68          & 71.96$\pm$0.89          \\
& A4                   & 100$\%$                                                                           & 100$\%$                                                                          & 86.44$\pm$0.22          & 85.35$\pm$0.21          & 86.13$\pm$0.17          & \textbf{88.43$\pm$0.21} & 86.59$\pm$0.20                       & \textbf{75.95$\pm$0.85} & 71.64$\pm$0.55          & 70.84$\pm$0.63          & 74.35$\pm$0.55          & 73.20$\pm$0.33          \\ \hline
\multirow{4}{*}{\textbf{Ours}} & A1                   & 100$\%$                                                                           & 100$\%$                                                                          & \textbf{89.34$\pm$0.07} & 87.55$\pm$0.08          & 88.18$\pm$0.11          & 87.92$\pm$0.06          & \textbf{88.25$\pm$0.08}              & \textbf{86.41$\pm$0.53} & 72.79$\pm$0.26          & 72.44$\pm$0.37          & 76.35$\pm$0.43          & \textbf{77.00$\pm$0.20} \\
& A2                   & 100$\%$                                                                           & 100$\%$                                                                          & 87.89$\pm$0.08          & \textbf{89.27$\pm$0.05} & 87.65$\pm$0.06          & 87.21$\pm$0.08          & 88.01$\pm$0.07                       & 75.51$\pm$0.89          & \textbf{76.56$\pm$1.08} & 72.42$\pm$0.96          & 73.30$\pm$0.93          & 74.45$\pm$0.48          \\
& A3                   & 100$\%$                                                                           & 100$\%$                                                                          & 88.06$\pm$0.07          & 87.40$\pm$0.04          & \textbf{89.35$\pm$0.08} & 87.63$\pm$0.06          & 88.11$\pm$0.06                       & 74.76$\pm$1.45          & 75.53$\pm$0.84          & \textbf{78.16$\pm$1.18} & 73.68$\pm$1.12          & 75.53$\pm$0.58          \\
& A4                   & 100$\%$                                                                           & 100$\%$                                                                          & 87.67$\pm$0.06          & 86.73$\pm$0.11          & 87.62$\pm$0.12          & \textbf{89.22$\pm$0.08} & 87.81$\pm$0.09                       & \textbf{76.92$\pm$2.49} & 71.97$\pm$1.98          & 71.61$\pm$2.10          & 75.87$\pm$1.88          & 74.09$\pm$1.06          \\ \Xhline{1.5pt}
\multirow{2}{*}{D-person}                       & All                  & 25$\%$                                                                            & 100$\%$                                                                          & 86.68$\pm$0.10          & 87.93$\pm$0.09          & 88.22$\pm$0.12          & 86.17$\pm$0.24          & 87.25$\pm$0.09                       & 80.19$\pm$0.49          & 74.75$\pm$0.27          & 73.41$\pm$0.45          & 75.09$\pm$0.54          & 75.86$\pm$0.42          \\
& All                  & 50$\%$                                                                            & \textbf{200$\%$}                                                                 & 87.53$\pm$0.09          & \textbf{88.62$\pm$0.09} & \textbf{88.82$\pm$0.14} & 86.75$\pm$0.26          & 87.93$\pm$0.11                       & 81.26$\pm$0.56          & 76.16$\pm$0.25          & 75.08$\pm$0.37          & 76.18$\pm$0.45          & 77.17$\pm$0.38          \\ \hline
\multirow{3}{*}{CE Loss}                        & Sample               & 25$\%$                                                                            & 25$\%$                                                                           & 83.46$\pm$0.59          & 82.67$\pm$0.48          & 83.34$\pm$0.48          & 82.44$\pm$0.54          & 82.98$\pm$0.52                       & 75.73$\pm$1.47          & 71.29$\pm$0.95          & 70.63$\pm$1.21          & 71.47$\pm$1.59          & 74.78$\pm$0.96          \\
& Sample               & 50$\%$                                                                            & 50$\%$                                                                           & 85.69$\pm$0.35          & 84.88$\pm$0.29          & 85.81$\pm$0.26          & 84.90$\pm$0.34          & 85.32$\pm$0.31                       & 76.63$\pm$1.66          & 73.30$\pm$0.85          & 72.13$\pm$1.53          & 73.09$\pm$1.40          & 75.29$\pm$0.91          \\
& Sample               & 100$\%$                                                                           & 100$\%$                                                                          & 87.23$\pm$0.18          & 86.81$\pm$0.22          & 86.89$\pm$0.12          & 86.36$\pm$0.18          & 86.82$\pm$0.18                       & 78.06$\pm$1.35          & 74.84$\pm$1.01          & 74.44$\pm$0.99          & 74.46$\pm$1.04          & 75.45$\pm$0.63          \\ \hline
\multirow{3}{*}{\textbf{Ours}} & Sample               & 25$\%$                                                                            & 25$\%$                                                                           & 86.41$\pm$0.16          & 85.77$\pm$0.26          & 86.37$\pm$0.30          & 86.00$\pm$0.21          & 86.14$\pm$0.23                       & 79.18$\pm$1.61          & 74.54$\pm$0.94          & 73.71$\pm$1.61          & 74.52$\pm$1.73          & 75.49$\pm$0.84          \\
& Sample               & 50$\%$                                                                            & 50$\%$                                                                           & 87.50$\pm$0.11          & 86.82$\pm$0.12          & 87.53$\pm$0.08          & 87.47$\pm$0.07          & 87.33$\pm$0.10                       & 79.69$\pm$1.40          & 75.83$\pm$0.96          & 74.56$\pm$1.58          & 75.70$\pm$1.52          & 76.45$\pm$0.76          \\
& Sample               & 100$\%$                                                                           & 100$\%$                                                                          & \textbf{88.40$\pm$0.11} & 87.83$\pm$0.15          & 88.07$\pm$0.15          & \textbf{88.04$\pm$0.09} & \textbf{88.09$\pm$0.13}              & \textbf{81.31$\pm$1.10} & \textbf{77.16$\pm$1.02} & \textbf{76.20$\pm$1.22} & \textbf{76.82$\pm$1.04} & \textbf{77.87$\pm$0.58} \\ \Xhline{1.5pt}
CE Loss                                         & Sample               & 100$\%$                                                                           & 100$\%$                                                                          & 87.23$\pm$0.18          & 86.81$\pm$0.22          & 86.89$\pm$0.12          & 86.36$\pm$0.18          & 86.82$\pm$0.18                       & 78.06$\pm$1.35          & 74.84$\pm$1.01          & 74.44$\pm$0.99          & 74.46$\pm$1.04          & 75.45$\pm$0.63          \\
RMD                                             & Sample               & 100$\%$                                                                           & 100$\%$                                                                          & 87.71$\pm$0.20          & 87.13$\pm$0.17          & 87.34$\pm$0.12          & 86.72$\pm$0.20          & 87.23$\pm$0.17                       & 80.30$\pm$0.70          & 76.39$\pm$0.49          & 75.15$\pm$0.88          & 75.67$\pm$0.65          & 76.88$\pm$0.68          \\
ADELE                                           & Sample               & 100$\%$                                                                           & 100$\%$                                                                          & 85.65$\pm$0.76          & 85.54$\pm$0.70          & 85.75$\pm$0.63          & 84.89$\pm$0.69          & 85.46$\pm$0.70                       & \textbf{79.97$\pm$1.23} & 76.15$\pm$0.65          & 75.10$\pm$0.77          & 75.10$\pm$1.23          & 76.58$\pm$0.65          \\
CDR                                             & Sample               & 100$\%$                                                                           & 100$\%$                                                                          & 87.01$\pm$0.38          & 86.74$\pm$0.32          & 86.62$\pm$0.28          & 85.98$\pm$0.34          & 86.59$\pm$0.33                       & 76.81$\pm$2.45          & 73.70$\pm$1.29          & 72.71$\pm$1.77          & 72.90$\pm$1.99          & 74.03$\pm$1.03          \\
Co-Teaching                                     & Sample               & 100$\%$                                                                           & 100$\%$                                                                          & 87.26$\pm$0.28          & 86.77$\pm$0.26          & 87.08$\pm$0.23          & 86.43$\pm$0.27          & 86.89$\pm$0.26                       & 79.85$\pm$1.57          & 75.35$\pm$0.68          & 75.10$\pm$0.94          & 75.40$\pm$1.37          & 76.43$\pm$0.70          \\
IDMPS                                           & Sample               & 100$\%$                                                                           & 100$\%$                                                                          & 87.62$\pm$0.15          & 87.33$\pm$0.25          & 87.18$\pm$0.29          & 87.18$\pm$0.17          & 87.33$\pm$0.22                       & 79.20$\pm$1.19          & 74.88$\pm$0.61          & 74.30$\pm$0.74          & 74.61$\pm$0.91          & 75.75$\pm$0.55          \\
JoCoR                                           & Sample               & 100$\%$                                                                           & 100$\%$                                                                          & 87.26$\pm$0.37          & 86.90$\pm$0.28          & 87.14$\pm$0.25          & 86.75$\pm$0.30          & 87.01$\pm$0.30                       & 79.23$\pm$2.01          & 75.65$\pm$0.55          & 74.56$\pm$1.37          & 75.13$\pm$1.56          & 76.14$\pm$0.84          \\
SP-Guide                                        & Sample               & 100$\%$                                                                           & 100$\%$                                                                          & 77.56$\pm$2.21          & 77.06$\pm$2.20          & 76.48$\pm$2.25          & 76.98$\pm$2.26          & 77.02$\pm$2.23                       & 49.51$\pm$1.50          & 46.57$\pm$1.69          & 47.66$\pm$1.49          & 49.07$\pm$1.41          & 48.20$\pm$0.77          \\
MSA                                                       & Sample               & 100$\%$                                                                           & 100$\%$                                                                          & 88.24$\pm$0.06                & 87.66$\pm$0.09            & 87.52$\pm$0.09            & 87.29$\pm$0.05            & {87.68$\pm$0.07}              & 75.48$\pm$0.37              & 72.40$\pm$0.26              & 71.74$\pm$0.18              & 71.55$\pm$0.31              & 72.79$\pm$0.28              \\
T-Loss                                                    & Sample               & 100$\%$                                                                           & 100$\%$                                                                          & {87.72$\pm$0.29}      & {87.27$\pm$0.21}     & {87.51$\pm$0.21}     & {87.24$\pm$0.27}     & {87.44$\pm$0.25}     & {78.28$\pm$1.16}     & {72.88$\pm$0.99}     & {72.48$\pm$1.07}     & {74.31$\pm$1.09}     & {74.49$\pm$1.08}     \\
\textbf{Ours}                                   & Sample               & 100$\%$                                                                           & 100$\%$                                                                          & \textbf{88.40$\pm$0.11} & \textbf{87.83$\pm$0.15} & \textbf{88.07$\pm$0.15} & \textbf{88.04$\pm$0.09} & \textbf{88.09$\pm$0.13}              & \textbf{81.31$\pm$1.10} & \textbf{77.16$\pm$1.02} & \textbf{76.20$\pm$1.22} & \textbf{76.82$\pm$1.04} & \textbf{77.87$\pm$0.58} \\ \Xhline{1.5pt}
\end{tabular}
}
\end{table*}
\begin{figure*}[!h]
	\centering
	\includegraphics[width=0.99\textwidth]{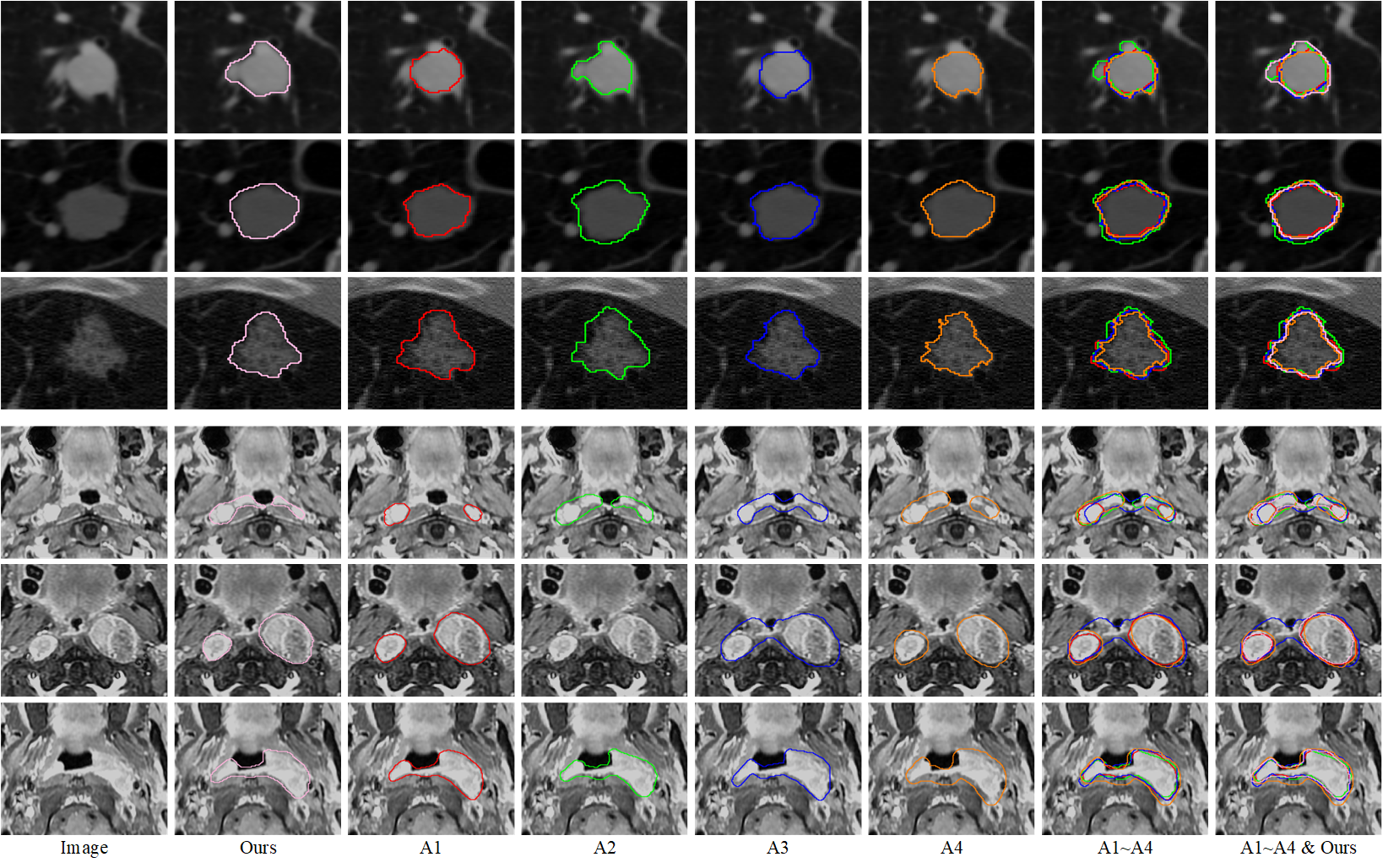}		
	\caption{Qualitative comparisons on the LIDC dataset (upper) and the MMIS-2024 dataset (lower). A1$\sim$A4 denote annotation from different experts.}
	\label{seg_visual_lidc_npc}
\end{figure*}
\subsubsection{Results on the LIDC and MMIS-2024 dataset}
Table \ref{LIDCMMIS} summarizes the quantitative performance of our proposed GSD-Net compared to several existing methods on the LIDC and MMIS-2024 datasets.
(1) U-Net models trained on a single expert’s annotations perform well when evaluated on that expert’s annotations but show notable performance drops on annotations from other experts, revealing subjectivity and inconsistency in labeling. In contrast, GSD-Net exhibits moderate degradation under cross-expert evaluation, indicating stronger robustness to inter-expert variability.
(2) To better reflect real-world multi-expert scenarios, we adopt a sampling strategy that selects one expert annotation per image during training. This introduces natural label variability and enables evaluation under realistic annotation noise. Under this setting, GSD-Net achieves an average Dice score of 88.09\% on the LIDC and 77.87\% on the MMIS-2024 dataset, outperforming all compared approaches under the same training conditions. 
(3) We further compare with D-Persona {{\citep{wu2024diversified}}}, a representative multi-rater method that uses all four expert annotations per image to generate personalized predictions. When trained on 50\% of images, which results in double the annotation cost, D-Persona achieves an average Dice score of 87.93\% on the LIDC dataset and 77.17\% on MMIS-2024 dataset, both lower than those achieved by GSD-Net. 

In addition to the quantitative analysis, the qualitative segmentation results in Fig.~\ref{seg_visual_lidc_npc} further highlight the strengths of GSD-Net, showing that it produces segmentations closer to the true semantic regions, with improved accuracy and more precise delineation of lesion boundaries.
Taken together, these findings demonstrate that GSD-Net effectively addresses inter-expert variability while maintaining stable performance across datasets, underscoring its potential for real-world clinical deployment in scenarios where label noise arises from inter-annotator variability.

\begin{figure*}[t]
	\centering
	\includegraphics[width=0.99\textwidth]{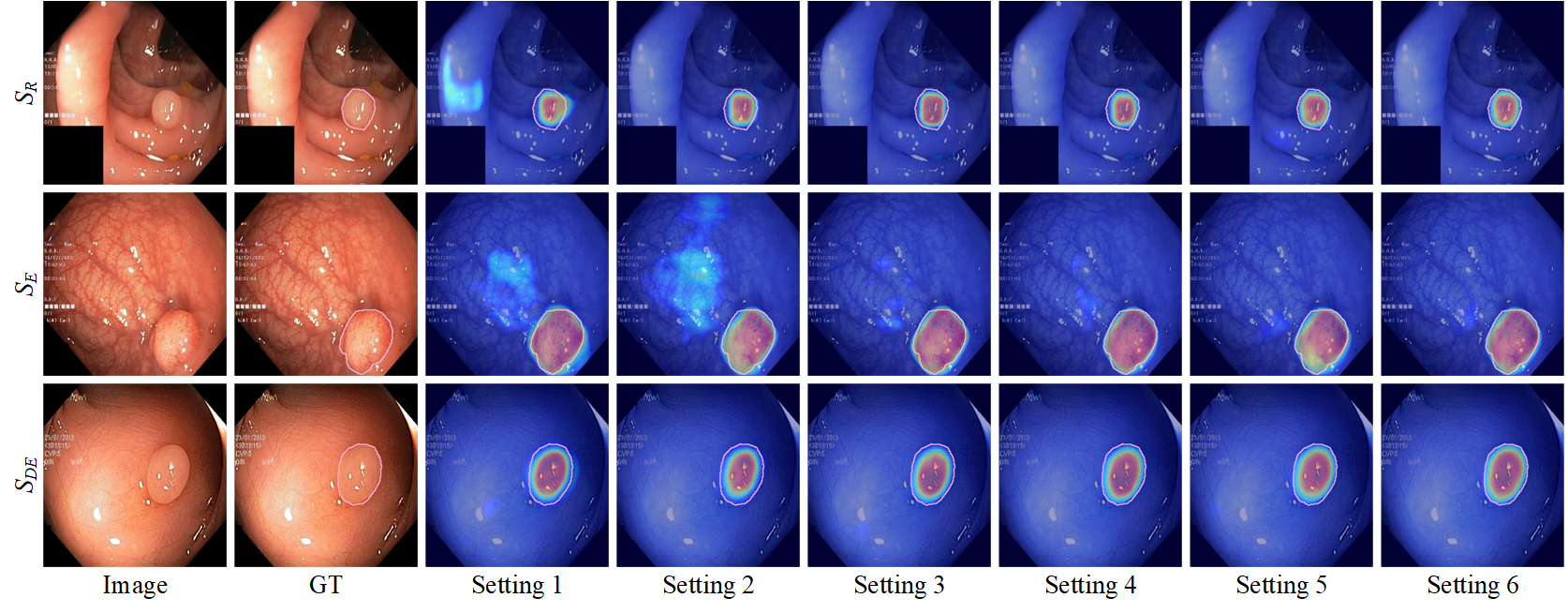}		
	\caption{Grad-CAM visualizations under different ablation settings (Settings 1-6), consistent with the configurations shown in Table~\ref{abu_module}. The pink contours denote the boundary of ground truth.}
	\label{abu_cam_figure}
\end{figure*}
\subsection{Ablation Study}
To evaluate the contribution of each component, we conducted ablation studies on the Kvasir dataset. Using $\mathcal{L}_{{JoCoR}}$ (Eq.~\eqref{loss_jocor}) as the baseline, we analyzed the impact of the GDA, SGLR, and KT modules. Detailed results are reported in Table \ref{abu_module}. The GDA module, which introduces geometric distance–aware reweighting, significantly improved segmentation accuracy across all three types, demonstrating its effectiveness to mitigate annotation noise. 
The SGLR module further enhances performance by incorporating superpixel-based structural priors to refine boundaries and correct labels, outperforming the LR variant. Building on these gains, the KT module delivers additional improvements by facilitating cross-sample knowledge transfer, which increased data diversity and reduced the influence of noisy labels. Together, these results demonstrate that each module is effective and complementary in enhancing robustness. Grad-CAM visualizations (Fig.~\ref{abu_cam_figure}) further illustrate that as modules are integrated, the model progressively focuses on relevant features, especially lesion boundaries, providing visual evidence of improved interpretability.

\begin{table}[t]
	\centering
	\caption{Ablation study of component combinations on the Kvasir dataset. LR indicates the label refinement module without superpixel processing, and SGLR denotes the version with structure-guide integration.}
	\label{abu_module}
	\resizebox{0.48\textwidth}{!}{
		\begin{tabular}{m{0.35cm}<{\centering}|m{0.8cm}<{\centering}m{0.5cm}<{\centering}m{0.3cm}<{\centering}m{0.6cm}<{\centering}m{0.4cm}<{\centering}|c|c|c}
			\hline
			Set. & $\mathcal{L}_{JoCoR}$     & GDA                       & LR                        & SGLR                      & KT                        & $S_R$          & $S_E$          & $S_{DE}$       \\ \hline
			1      & \checkmark &                           &                           &                           &                           & 71.05$\pm$0.78 & 73.79$\pm$0.79 & 69.23$\pm$1.15 \\
			2      & \checkmark & \checkmark &                           &                           &                           & 76.76$\pm$0.64 & 75.91$\pm$0.37 & 75.13$\pm$0.48 \\
			3      & \checkmark & \checkmark & \checkmark &                           &                           & 78.74$\pm$0.41 & 76.73$\pm$0.29 & 77.26$\pm$0.60 \\
			4      & \checkmark & \checkmark & \checkmark & \checkmark &                           & 78.84$\pm$0.46 & 78.94$\pm$0.62 & 78.56$\pm$0.53 \\
			5      & \checkmark & \checkmark & \checkmark &                           & \checkmark & 78.94$\pm$0.35 & 78.47$\pm$0.46 & 78.78$\pm$0.35 \\
			6      & \checkmark & \checkmark & \checkmark & \checkmark & \checkmark & \textbf{80.04$\pm$0.42} & \textbf{79.39$\pm$0.33} & \textbf{79.97$\pm$0.53} \\ \hline
		\end{tabular} 
	}
\end{table}

\section{Discussion}
In medical image segmentation, label noise is often unavoidable due to intra- and inter-observer variability, indistinct lesion boundaries, and coarse delineations that arise when lesions are annotated in a rough or imprecise manner (Fig.~\ref{show_question}), which can impair the network’s ability to capture object features. To address these challenges, we propose GSD-Net, a framework designed to enhance robustness under noisy annotations. To validate its effectiveness, we first validated the framework on four public datasets with simulated noise, employing not only the widely used morphology-based erosion and dilation strategy ($S_{DE}$) but also Markov-based boundary perturbation strategies ($S_{R}$, $S_{E}$) to more accurately simulate the irregular boundaries observed in clinical practice (Fig. \ref{show_simulated_noise}). 
Moreover, in large-scale medical datasets, collaborative annotations from multiple experts make inter-observer variability a frequent and realistic source of label noise. Results on the LIDC and MMIS-2024 datasets (Table~\ref{LIDCMMIS} and Fig.~\ref{seg_visual_lidc_npc}) highlight this challenge: conventional models exhibit notable performance drops under cross-expert evaluation, whereas GSD-Net maintains stable performance. These findings demonstrate that our framework more effectively addresses inter-observer variability than existing methods, thereby improving its reliability in real-world multi-rater scenarios. These results underscore the clinical value of GSD-Net, as its robustness to inter-observer variability enables more consistent and reliable segmentation across different radiologists, thereby facilitating reproducible diagnosis and treatment planning.

\begin{table*}[]
\centering
	\caption{{Performance of GSD-Net integrated with UNet++, ViT U-Net and U-Net under two training strategies: GSD-Net and CE Loss.}}
    \label{abu_backbone}
\resizebox{1\textwidth}{!}{
    \begin{tabular}{lccccccccc}
    \hline
    \multicolumn{10}{c}{GSD-Net} \\ \hline
    \multicolumn{1}{l|}{Backbones} & \multicolumn{3}{c|}{Kvasir Dataset}                                   & \multicolumn{3}{c|}{Shenzhen Dataset}                                 & \multicolumn{3}{c}{BU\_SUC Dataset}              \\ \hline
    \multicolumn{1}{l|}{UNet++}    & 78.95$\pm$0.60 & 78.78$\pm$0.50 & \multicolumn{1}{c|}{78.67$\pm$0.39} & 93.84$\pm$0.06 & 90.58$\pm$0.16 & \multicolumn{1}{c|}{94.86$\pm$0.57} & 77.33$\pm$2.43 & 84.00$\pm$0.41 & 89.68$\pm$0.16 \\
    \multicolumn{1}{l|}{ViT U-Net} & 82.37$\pm$0.38 & 80.68$\pm$0.30 & \multicolumn{1}{c|}{82.65$\pm$0.28} & 93.86$\pm$0.08 & 90.14$\pm$0.17 & \multicolumn{1}{c|}{94.45$\pm$0.10} & 83.28$\pm$0.63 & 86.21$\pm$0.21 & 90.06$\pm$0.09 \\
    \multicolumn{1}{l|}{U-Net}     & 80.04$\pm$0.42 & 79.39$\pm$0.33 & \multicolumn{1}{c|}{79.97$\pm$0.53} & 93.31$\pm$0.09 & 89.25$\pm$0.25 & \multicolumn{1}{c|}{94.68$\pm$0.07} & 80.27$\pm$0.68 & 84.59$\pm$0.41 & 89.30$\pm$0.25 \\ \hline
    \multicolumn{10}{c}{CE Loss}  \\ \hline
    \multicolumn{1}{l|}{Backbones} & \multicolumn{3}{c|}{Kvasir Dataset}                                   & \multicolumn{3}{c|}{Shenzhen Dataset}                                 & \multicolumn{3}{c}{BU\_SUC Dataset}              \\ \hline
    \multicolumn{1}{l|}{UNet++}    & 60.26$\pm$3.16 & 65.58$\pm$2.44 & \multicolumn{1}{c|}{58.12$\pm$3.15} & 61.95$\pm$1.10 & 73.90$\pm$1.76 & \multicolumn{1}{c|}{77.11$\pm$3.22} & 58.94$\pm$1.56 & 73.41$\pm$0.80 & 79.90$\pm$0.68 \\
    \multicolumn{1}{l|}{ViT U-Net} & 69.21$\pm$0.51 & 72.06$\pm$0.56 & \multicolumn{1}{c|}{69.78$\pm$0.28} & 60.63$\pm$0.86 & 74.39$\pm$0.16 & \multicolumn{1}{c|}{85.08$\pm$0.43} & 60.58$\pm$1.14 & 74.75$\pm$0.31 & 84.65$\pm$0.86 \\
    \multicolumn{1}{l|}{U-Net}     & 66.86$\pm$4.46 & 70.33$\pm$3.25 & \multicolumn{1}{c|}{63.11$\pm$2.54} & 60.36$\pm$1.68 & 73.82$\pm$1.58 & \multicolumn{1}{c|}{76.08$\pm$3.01} & 58.14$\pm$2.75 & 73.01$\pm$1.52 & 79.54$\pm$2.99 \\ \hline
    \end{tabular} }
\end{table*}

Extensive experiments across six datasets spanning endoscopy, X-ray, US, MRI, and CT modalities show that GSD-Net achieves superior performance under diverse noise conditions, demonstrating strong generalizability and stability across different modalities and noise types. Furthermore, Grad-CAM visualizations (Fig.~\ref{abu_cam_figure}) provide qualitative evidence that GSD-Net progressively focuses on more relevant and anatomically meaningful structures, supporting its interpretability.

Another notable strength is that GSD-Net serves as a training-level enhancement independent of specific network architectures, ensuring broad applicability and generalization. To further validate this property, we evaluated the framework on two alternative backbones: UNet++ {{\citep{unetpp}}} and a U-Net variant with a Mix Vision Transformer encoder (ViT U-Net) {{\citep{xie2021segformer}}}. As shown in Table~\ref{abu_backbone}, GSD-Net consistently improved segmentation performance for both backbones, underscoring its robustness and adaptability across diverse segmentation models.

{Furthermore, in light of the recent emergence of medical foundation models, we extended our comparative analysis to include the Medical SAM Adapter (MSA) {\citep{wu2025medical}}. While models like SAM provide powerful spatial priors and exceptional feature representations, our results (see Table {\ref{table_Kvasir_shenzhen_busuc}} and {\ref{LIDCMMIS}}) indicate that foundational priors alone cannot inherently circumvent the fundamental challenge of label noise. Without an explicit noise-handling mechanism, even these large-scale models remain susceptible to misleading supervision. This observation underscores the necessity of our GSD-Net, which can potentially be integrated with foundation models during the fine-tuning phase to achieve a synergy between superior generalization and extreme noise tolerance. 
Due to its architecture-agnostic nature, our framework is uniquely positioned to potentially function as a robust filtering layer during the domain-specific fine-tuning of foundation models. Such an integration holds the promise of ensuring that the inherent structural integrity of large-scale priors is not compromised by noisy or low-quality expert annotations. This synergy defines a promising research trajectory, establishing our noise-suppression strategy as a pivotal component for developing reliable medical AI systems in the burgeoning era of foundation models.}

{Beyond empirical robustness, an important design consideration of GSD-Net lies in how label denoising is regularized under noisy supervision. The KT module is therefore designed as a structural regularization component, rather than a consistency-based semi-supervised learning mechanism. By introducing controlled cross-sample structural perturbations through local region exchange, KT exposes the model to more diverse structural contexts during training. This encourages the model to rely less on individual noisy annotations and instead focus on anatomical patterns that remain stable across varying structural configurations, thereby improving sensitivity to fine-grained and noise-obscured boundaries and enabling more reliable recovery of corrupted labels. Unlike semi-supervised approaches that enforce prediction consistency across different views of the same sample, KT deliberately leverages structural discrepancies across samples as a regularization signal to enhance robustness under noisy supervision.}

{It is worth noting that several components of the proposed framework are related to existing techniques in noisy-robust learning, and part of these strategies have been developed for segmentation tasks. However, many were originally designed for classification under the assumption of independently distributed label noise. In contrast, annotation errors in medical image segmentation often exhibit strong spatial structure, with noise concentrated around anatomical boundaries due to contour ambiguity and inter-observer variability. Motivated by this, our framework reformulates and integrates these strategies within a unified collaborative learning paradigm. Spatially-aware supervision, dynamic superpixel-guided label refinement, and knowledge transfer are jointly optimized to explicitly handle spatially structured noise, thereby improving robustness to noise annotations. Therefore, the contribution lies not only in the individual components; but more importantly in their principled redesign and synergistic integration for robust medical image segmentation under structured label noise.}
\begin{figure}[!h]
	\centering
	\includegraphics[width=0.48\textwidth]{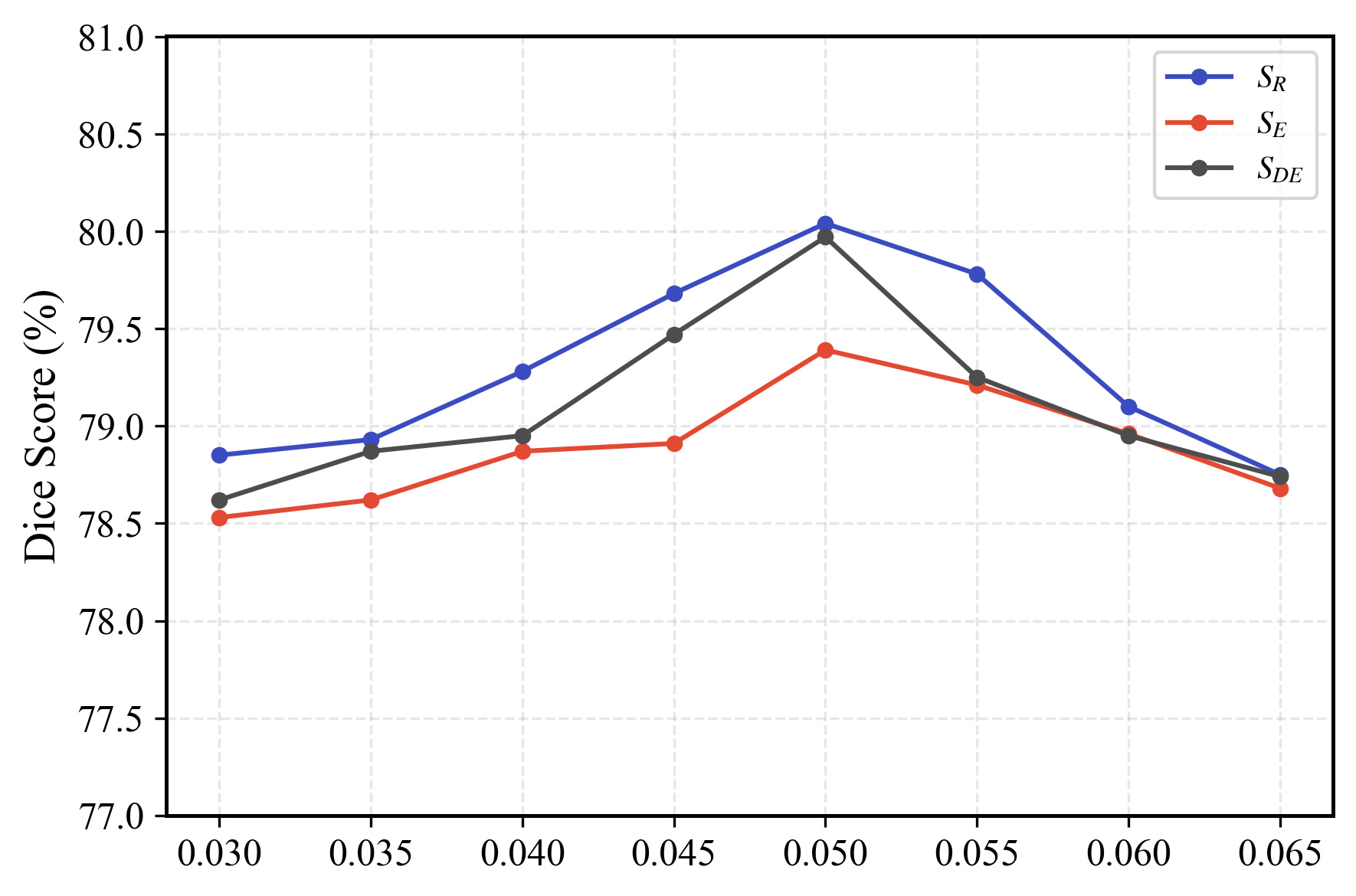}		
	\caption{Effect of hyperparameters $\tau$ on segmentation performance on the Kvasir dataset.}
	\label{abu_tau_sensitivity}
\end{figure}

Additionally, GSD-Net exhibits limited sensitivity to the hyperparameter $\tau$, which is introduced in the small-loss strategy to filter noisy samples. As demonstrated in the ablation study on the Kvasir dataset (Fig.~\ref{abu_tau_sensitivity}), variations in $\tau$ led to minor performance changes, and the model maintained strong segmentation accuracy even under suboptimal values. These results highlight the framework’s robustness to hyperparameter selection and its reliability under imperfect noise-level estimation. 
\begin{figure*}[!h]
	\centering   
	\includegraphics[width=1\textwidth]{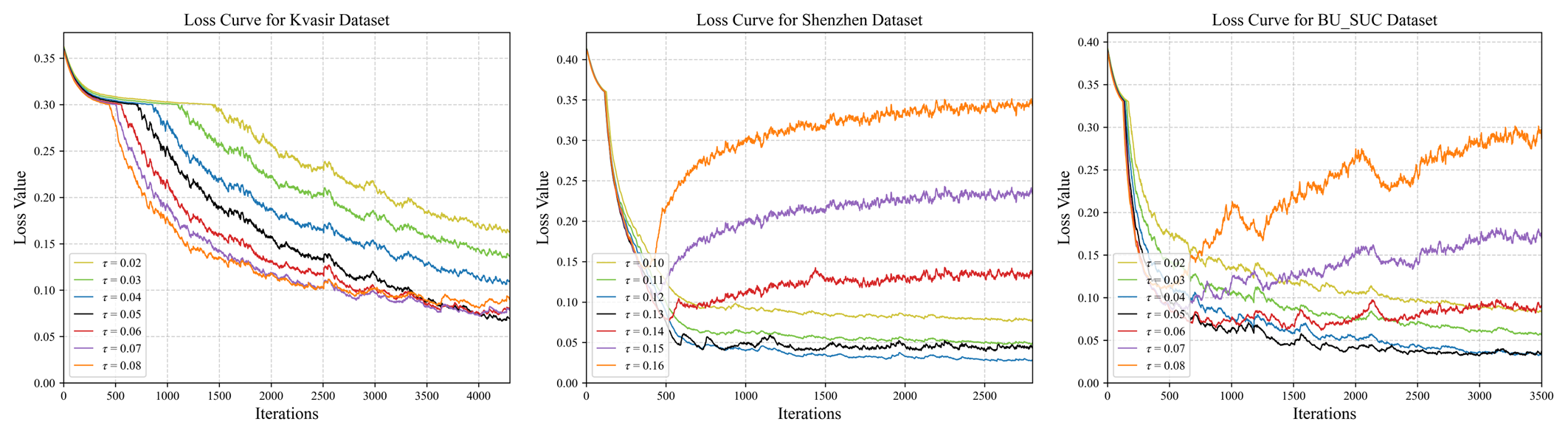}   
	\caption{{Sensitivity analysis of $\mathcal{L}_{search}$ curves. Loss trajectories under varying drop-rates $\tau$ on three datasets with distinct ground-truth noise rates: 5.02\% (Kvasir), 12.33\% (Shenzhen), and 4.43\% (BU\_SUC). Each trajectory achieves its local minimum when $\tau$ closely approximates the intrinsic label noise density of the respective dataset.}}
	\label{fig:3_li_2} 
\end{figure*}
{To further refine this selection process in practical applications, we developed an architecture-facilitated search mechanism that leverages the intrinsic learning dynamics of GSD-Net. Central to this approach is a specialized loss function, $\mathcal{L}_{search}$, which enables the model to autonomously identify the optimal $\tau$ even in the absence of prior noise knowledge. Given the separation of $\mathbb{D}^{clean}$ defined in Eq. {\eqref{D_clean}} (Sec. {\ref{method_jocor}}), the search loss can be formulated as follows:}
\begin{align}
            \displaystyle
			\mathcal{L}_{search}= \frac{\sum_{(i,j) \in \mathbb{D}^{clean}} {\mathcal{L}_{sup}(x, nGT)} + \sum_{(i,j) \in \overline{\mathbb{D}^{clean}}} {\mathcal{L}_{sup}(x, \overline{nGT})}}{|\mathbb{D}|} ,
\end{align}
{where $\mathbb{D}^{clean}$ and $\overline{\mathbb{D}^{clean}}$ denote the identified clean regions and their complement based on the small-loss strategy, while $nGT$ and $\overline{nGT}$ represent the noisy labels and their inverted counterparts, respectively. 
By monitoring the evolutionary trajectories of $\mathcal{L}_{search}$, we empirically observed that the loss reaches its global minimum when $\tau$ closely aligns with the intrinsic noise rate of the dataset (Fig. {\ref{fig:3_li_2}}). This mechanism effectively transmutes $\tau$ from a manually tuned hyperparameter into a model-searchable variable, providing a robust and automated reference for practical applications where precise noise estimation is challenging. }
\begin{figure*}[!h]
	\centering
	\includegraphics[width=1\textwidth]{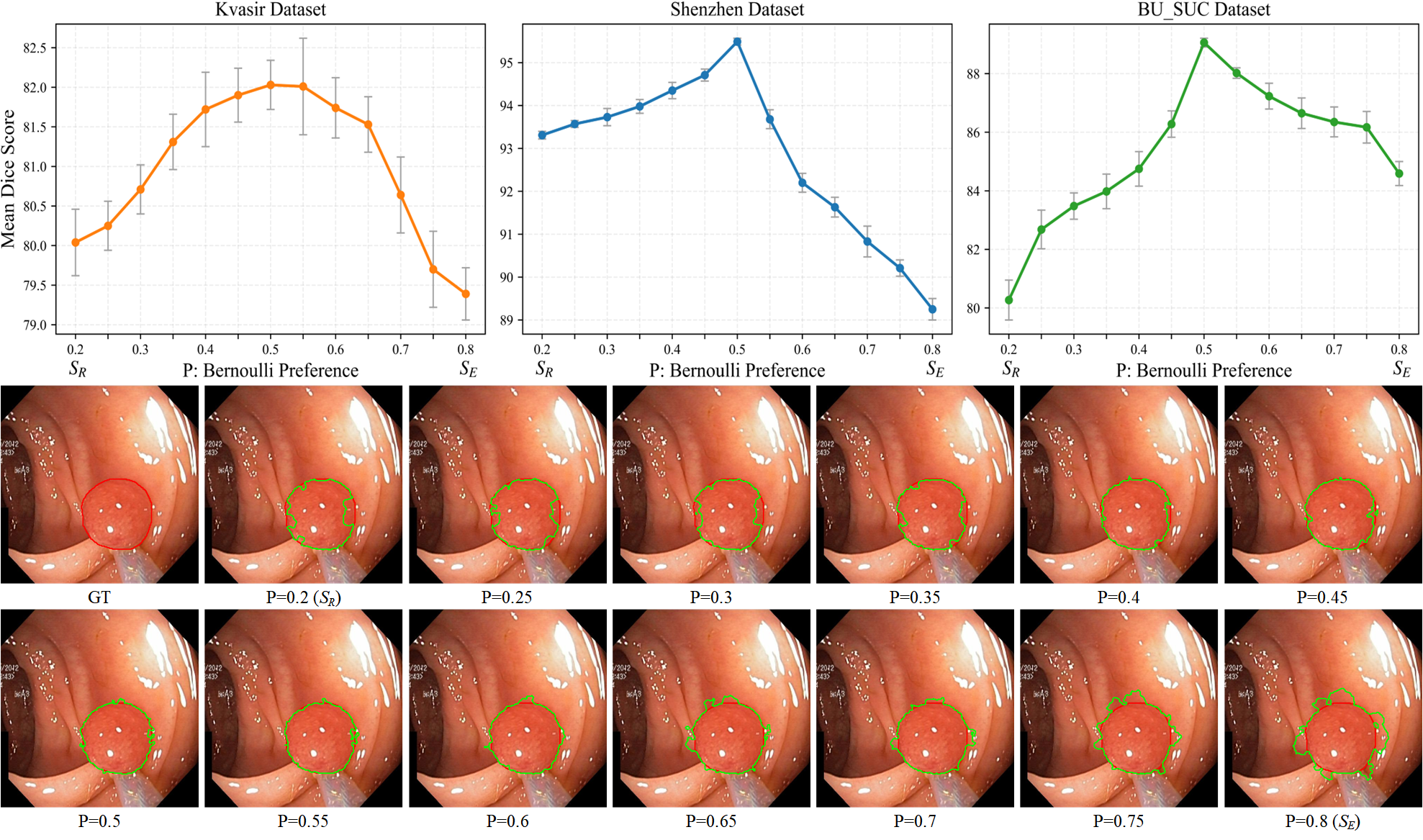}	
	\caption{{Segmentation results under varying noise intensities (upper panel) and corresponding visualization of the noise parameter P (Bernoulli Preference, see Table {\ref{hyperparameters_simulated}} (lower panel); ranges from 0.2 to 0.8, midpoint 0.5; smaller values indicate greater foreground reduction, larger values indicate greater foreground increase).}}
	\label{show_different_noise_type}
\end{figure*}

{
Similarly, the weight clipping threshold $T$ is a key factor in the GDA module, balancing robustness and gradient stability. A moderate $T$ effectively suppresses annotation noise, which is typically concentrated near object boundaries, while avoiding over-weighting distant background regions that could destabilize training. In practical applications, $T$ can be adapted to image resolution and the scale of target structures. For instance, smaller structures such as nuclei or vessels may require a lower $T$, whereas larger structures can tolerate higher values. This flexibility allows the GDA module to maintain stable performance across diverse imaging scenarios. Furthermore, for the SLIC superpixel method, the parameter $n_segments$ should also be adaptively adjusted according to the input image resolution. For example, for an image of size $512 \times 512$, $n_{segments}$ can be set to 4096.
}

{In addition to its flexible hyperparameter design, GSD-Net demonstrates superior robustness across a broad spectrum of noise intensities. As illustrated in Fig. {\ref{show_different_noise_type}}, the proposed framework maintains high segmentation fidelity even under severe label corruption, while exhibiting a progressive performance improvement as the noise intensity decreases. This trend confirms that GSD-Net remains highly responsive to increasing label quality, ensuring its consistent applicability across diverse clinical annotation qualities.}

\begin{table}[!h]
    \centering
    \caption{Performance across different proportions of clean labels trained with U-Net, and $S_{DE}$ noisy labels trained with GSD-Net.}
    \label{tab:noise_results}
    \resizebox{0.48\textwidth}{!}{
    \begin{tabular}{l|cccc|c}
    \hline
    Annotation  & \multicolumn{4}{c|}{GT (Clean Labels)}                            & $S_{DE}$ (Noisy) \\ \hline
    Ratio    & 100\%          & 50\%           & 25\%           & 12.5\%         & 100\%            \\ \hline
    Kvasir   & \textbf{80.62$\pm$1.39} & 75.99$\pm$2.28 & 68.82$\pm$1.28 & 55.93$\pm$4.04 & 79.97$\pm$0.53   \\
    Shenzhen & \textbf{95.24$\pm$0.30} & 94.93$\pm$0.39 & 94.57$\pm$0.28 & 94.20$\pm$0.58 & 94.68$\pm$0.07   \\
    BU\_SUC  & \textbf{89.95$\pm$1.00} & 88.26$\pm$0.78 & 87.40$\pm$0.74 & 83.57$\pm$1.35 & 89.30$\pm$0.25   \\ \hline
    \end{tabular} }
\end{table}
In practical applications, GSD-Net shows strong potential for reducing annotation costs. Annotations produced by junior doctors or medical students with limited training may contain noise. However, when processed through the proposed framework, they can still yield competitive performance. As shown in Table~\ref{tab:noise_results}, under the mild noise setting ($S_{DE}$), the model achieves results comparable to those obtained with fully clean labels. In contrast, reducing the proportion of clean annotations to 12.5\% results in a substantial performance drop on some datasets. These results suggest that a small amount of expert annotation alone is insufficient, whereas larger volumes of coarse annotations from less experienced annotators, combined with noise-robust learning, offer greater potential for competitive performance. Overall, the results underscore the importance of striking a balance between annotation precision and dataset scale, thereby mitigating reliance on exhaustive expert labeling and leveraging coarse delineations to improve efficiency while maintaining reliable segmentation under noisy supervision.

{From a deployment perspective, GSD-Net strikes an balance between segmentation efficacy and computational overhead. As illustrated in Table {\ref{compare_times}}, while the collaborative modules introduce a marginal increase in training-phase resource consumption, this overhead is well-justified by the substantial gains in model robustness and accuracy. Considering the rapid advancement of modern GPU hardware, such training costs remain entirely within a feasible and acceptable range for practical research and clinical development. Crucially, the framework maintains a highly efficient inference latency of 5.12 ms per image, confirming its capability to handle high-throughput, real-time medical image analysis in demanding clinical environments.}
\begin{table}[!h]
\centering
	\caption{{Comparison of computational cost of different methods on Kvasir, Shenzhen, and BU\_SUC datasets}}
    \label{compare_times}
\resizebox{0.5\textwidth}{!}{
	\begin{tabular}{lccccc}
		\hline
		\multicolumn{1}{c}{\multirow{2}{*}{Method}} & \multicolumn{3}{c}{Training Time (min)} & \multirow{2}{*}{\makecell{Max GPU \\ Memory (G)}} & \multirow{2}{*}{\makecell{Inference Time \\ (ms/img)}} \\ \cline{2-4}
		\multicolumn{1}{c}{}                        & Kvasir      & Shenzhen      & BU\_SUC     &                                 &                           \\ \hline
		CE                                          & 4.42        & 3.39          & 3.93      & 1.67                            & 3.26                      \\
		GCE                                         & 4.21        & 3.08          & 3.69      & 1.67                            & 3.26                      \\
		RCE                                         & {4.11}        & {3.06}          & {3.32}      & {1.67}                            & {3.26}                      \\
		RMD                                         & 86.49       & 56.86         & 70.62     & 9.03                            & 5.12                      \\
		ADELE                                       & 14.28       & 9.29          & 11.95     & 12.83                           & 3.26                      \\
		CDR                                         & 4.14        & 3.04          & 3.65      & 1.68                            & 3.26                      \\
		Co-Teaching                                  & 6.06        & 4.12          & 5.08      & 2.92                            & 5.12                      \\
		IDMPS                                       & 9.06        & 6.05          & 7.42      & 3.11                            & 5.12                      \\
		Jocor                                       & 6.2         & 5.01          & 5.73      & 2.98                            & 5.12                      \\
		SP-guide                                   & 19.51       & 13.44         & 16.24     & 8.79                            & 5.12                      \\
		MSA                                         & 9.25        & 6.54          & 8.02      & 1.85                            & 20.8                      \\
		T-loss                                      & 6.43        & 4.36          & 5.34      & 1.68                            & 3.26                      \\
		Ours                                        & 14.41       & 9.43          & 11.98     & 5.65                            & 5.12                      \\ \hline
	\end{tabular}
    }
\end{table}

{Despite the effectiveness of the proposed design, several limitations warrant further investigation. First, the generalizability of the framework requires rigorous validation across multi-center clinical settings with diverse imaging protocols. Second, the current evaluation focuses on binary segmentation; extending the framework to multi-class scenarios remains challenging due to the limited scalability of existing noise simulation strategies in handling prevalent inter-class overlaps. 
Furthermore, while the GDA module excels at segmenting blob-like structures, its performance may diminish when encountering thin, elongated, or branching morphologies. 
To mitigate these issues and further enhance the clinical utility of our framework, we propose the following directions for future research:
}
\begin{itemize}
    \item {Topological and Multi-Scale Modeling: We are going to integrate graph-based architectures, such as Graph Convolutional Networks, with multi-scale extractors to capture long-range dependencies and fine-grained details. This integration may facilitate the preservation of structural continuity in complex regions (e.g., vasculature) and enhance robustness against scale variations.}
    
    \item {Multimodal Learning and Interpretability: Looking forward, the integration of multimodal cues represents a transformative trajectory for medical image analysis. A promising direction involves leveraging advanced vision-language pre-training protocols, such as the CLIP text encoder {\citep{radford2021learning}}, to establish a unified embedding space for medical imagery and radiology reports {\citep{li2025openvocabct}}. By endowing vision models with such rich semantic priors, we can effectively compensate for the inherent limitations of pixel-based representations. Future research may further expand this noise-robust paradigm, fostering the distillation of high-level conceptual features that remain latent within the visual modality alone. Such mechanisms are poised to become instrumental in navigating the challenges of ill-defined boundaries and stochastic noise. Ultimately, by catalyzing a paradigm shift toward evidence-based artificial intelligence, this direction will reinforce clinical trust and ensure the procedural transparency essential for next-generation automated diagnostic systems.}

    \item {Characterization and Clinical Value of Imaging Noise:
    In medical imaging, diagnostic interpretation is often compromised by image noise, motion blur, and low-contrast regions. Consequently, a rigorous analysis of noise sources and their underlying etiologies is of paramount importance. For instance, signal degradation in Magnetic Resonance Imaging (MRI) often stems from motion artifacts; identifying such patterns provides critical feedback for manufacturers to optimize hardware, such as refining patient stabilization systems to minimize kinetic interference. By tracing noise characteristics back to their physical or physiological origins, research can inform the iterative optimization of imaging instrumentation. Furthermore, these regions of high variability may harbor significant biological information pertinent to disease progression. Future research paradigms should transition from traditional morphological segmentation toward biological interpretation. By systematically investigating the latent correlations between the morphological features of boundary zones and clinical indicators, such as molecular biomarkers, the utility of automated methodologies in personalized diagnosis and prognostic assessment can be substantially enhanced.}
    \item {Multi-Center Clinical Validation: Finally, efforts will be directed toward developing large-scale, multi-class, and multi-center datasets with expert-consensus annotations. Such validation is crucial to verify the framework's generalizability across diverse imaging protocols and to ensure its robustness in real-world clinical environments.}
\end{itemize}



\section{Conclusion}
In this study, we proposed GSD-Net, a noise-robust framework for medical image segmentation. Building on the small-loss strategy and integrating geometric distance–aware weighting, structure-guided label refinement, and knowledge transfer mechanisms, the framework effectively suppresses label noise and improves segmentation reliability. Extensive experiments on six public datasets showed consistent gains over state-of-the-art methods under both simulated and real-world noise, underscoring the robustness and practical potential of GSD-Net for accurate segmentation under imperfect supervision.

\section{Acknowledgments}
This work was supported in part by National Natural Science Foundation of China under Grant 62171133; in part by Fuzhou University Outstanding Students Overseas Study Fund, in part by the ERC IMI under grant 101005122, the H2020 under Grant 952172; in part by MRC under Grant MC/PC/21013; in part by the Royal Society under Grant IEC\textbackslash NSFC\textbackslash 211235, in part by the NVIDIA Academic Hardware Grant Program, in part by the SABER project supported by Boehringer Ingelheim Ltd, in part by the NIHR Imperial Biomedical Research Centre under Grant RDA01; in part by the Wellcome Leap Dynamic Resilience, in part by the UKRI guarantee funding for Horizon Europe MSCA Postdoctoral Fellowships under Grant EP/Z002206/1; in part by the UKRI MRC Research Grant, TFS Research Grant MR/U506710/1; and in part by the UKRI Future Leaders Fellowship under Grant MR/V023799/1.
\bibliographystyle{model2-names.bst}\biboptions{authoryear}
\bibliography{reference_label_noise}

%

\end{document}